\documentclass[10pt]{TEAI}
\usepackage[utf8]{inputenc}
\usepackage{float}
\usepackage{tabularx}
\usepackage{colortbl}
\usepackage{url}
\usepackage[bottom]{footmisc}
\newcommand{\method}{\textsc{cst}}
\newcommand{\methodfirst}{\textbf{\method}}
\newcommand{\keyresult}[1]{\textcolor[HTML]{0066CC}{#1}}
\providecommand{\Description}[1]{}
\providecommand{\cstIconPath}{figures/assets/table_icons/}
\definecolor{CSTTableHeader}{HTML}{E8EFF7}
\definecolor{CSTTableStripe}{HTML}{F5F8FB}
\definecolor{CSTTableFocus}{HTML}{EAF1F8}
\definecolor{CSTSrcsccd}{HTML}{2364A5}
\definecolor{CSTSrcconda}{HTML}{7042AD}
\definecolor{CSTSrctoxicn}{HTML}{A85B00}
\definecolor{CSTSrcscamspot}{HTML}{B14A3E}
\definecolor{CSTSrchatebr}{HTML}{087C78}
\definecolor{CSTSeloriginal}{HTML}{536B86}
\definecolor{CSTSelrisk}{HTML}{B35C09}
\definecolor{CSTSelmmr}{HTML}{087C78}
\definecolor{CSTSelgroup}{HTML}{7042AD}
\definecolor{CSTEffectPositive}{HTML}{1765A5}
\definecolor{CSTEffectNegative}{HTML}{9A3F72}
\definecolor{CSTEffectNeutral}{HTML}{566170}
\definecolor{CSTEffectPositiveBg}{HTML}{EAF3FC}
\definecolor{CSTEffectNegativeBg}{HTML}{F8ECF3}
\definecolor{CSTEffectNeutralBg}{HTML}{F0F3F6}
\newcommand{\cstTableIcon}[1]{%
  \raisebox{-0.8mm}{\includegraphics[width=4mm,height=4mm,keepaspectratio]{\cstIconPath#1.png}}\hspace{0.6mm}}
\newcommand{\cstSource}[2]{\cstTableIcon{source_#1}\textcolor{CSTSrc#1}{#2}}
\newcommand{\cstSelector}[2]{\cstTableIcon{selector_#1}\textcolor{CSTSel#1}{#2}}
\newcommand{\cstNumberChip}[3]{\begingroup\setlength{\fboxsep}{0pt}%
  \colorbox{#1}{\textcolor{#2}{\hspace{1.2pt}\strut#3\hspace{1.2pt}}}\endgroup}
\newcommand{\cstDeltaPos}[1]{\cstNumberChip{CSTEffectPositiveBg}{CSTEffectPositive}{#1}}

\newcommand{\cstDeltaZero}[1]{\cstNumberChip{CSTEffectNeutralBg}{CSTEffectNeutral}{#1}}

\hypersetup{
  pdftitle={CST: Lightweight Risk Triage with Source-Grounded Evidence for Web Discussions},
  pdfauthor={Zhihua Wang; Yifeng Gao; Yunhao Feng; Rui Pan; Baiheng Zhao; Zixing Chen; Nan Li},
  pdfsubject={Risk triage and evidence organization for Web discussions}
}
\title{\mbox{\raisebox{-0.15em}[0pt][0pt]{\includegraphics[height=1.05em]{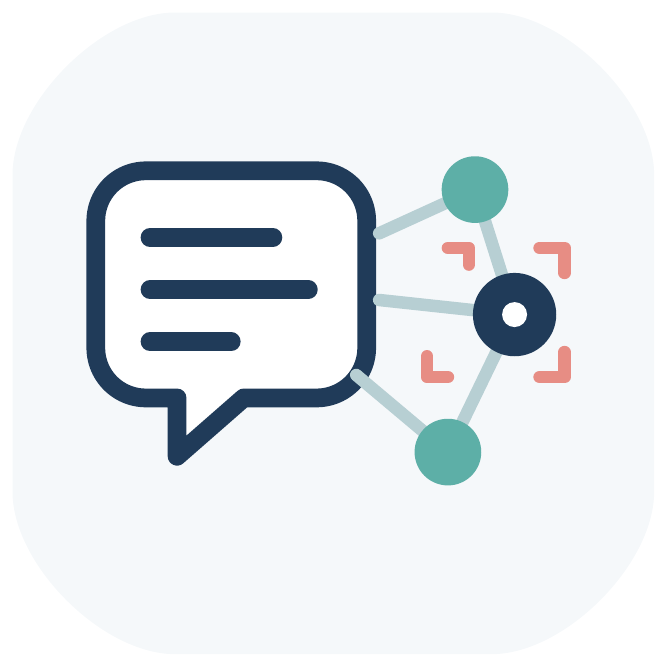}}\hspace{0.25em}CST:} Lightweight Risk Triage with Source-Grounded Evidence for Web Discussions}
\author{\mbox{Zhihua Wang\textsuperscript{1,2,*}},
  \mbox{Yifeng Gao\textsuperscript{1,$\dagger$}},
  \mbox{Yunhao Feng\textsuperscript{3}},\\[2pt]
  \mbox{Rui Pan\textsuperscript{1}},
  \mbox{Baiheng Zhao\textsuperscript{1}},
  \mbox{Zixing Chen\textsuperscript{1}},
  \mbox{Nan Li\textsuperscript{1}}}
\affiliation[1]{\mbox{Fudan University}}
\affiliation[2]{\mbox{University of Shanghai for Science and Technology}}
\affiliation[3]{\mbox{Alibaba Group}}
\abstract{Web communities rely on moderators to catch harmful messages such as insults, hate speech, and scams, but no one can read every message. Existing tools give each message a risk score, which tells moderators where to look but not what to read or how to check it. Yet meaning depends on context: the same reply, ``I agree.'', can support or oppose removing someone from a group. Moderators therefore need source-grounded evidence, not just scores: a few key messages kept with their original wording and replies. We present \methodfirst{}, a lightweight triage pipeline that first decides where to look, then what to show. It scores risk by pairing character patterns, which capture exact wording, with sentence meaning from a fixed multilingual model, so it handles several languages and runs on an ordinary CPU. It then groups related messages and picks from each group in turn, so a small reading budget covers different parts of a discussion, and links every pick back to the original thread. Across six tasks from five datasets in several languages, \method{} ranks harmful messages more accurately than a fully fine-tuned multilingual model and, under the same budget, finds more of them than reading by score alone. In a user study, reviewers needed at least 27\% less time with \method{}'s grouped evidence while keeping decision accuracy above 94\%. \method{} turns risk scores into evidence that moderators can read quickly and check in context. Code and dataset reconstruction resources will be released at \url{https://github.com/Kirito14IT/CST-MIL}.
}
\correspondence{\email{yifenggao23@m.fudan.edu.cn}}
\checkdata[Keywords]{Web discussions; risk triage; evidence selection; short text; topic organization; multilingual classification}

\begin{document}
\maketitle
\begingroup
\renewcommand{\thefootnote}{\fnsymbol{footnote}}
\footnotetext[1]{This work was completed during Zhihua Wang's internship at Fudan University.}
\footnotetext[2]{Corresponding author}
\endgroup
\vspace{-2\baselineskip}
\raggedbottom
\section{Introduction}
Web communities rely on moderators to keep discussions safe, and the workload is heavy. In a 2020 survey, 41\% of U.S. adults reported having personally experienced online harassment~\cite{vogels2021harassment}. On Reddit alone, volunteer moderators worked at least 466 hours per day in 2020, labor valued at about US\$3.4 million per year~\cite{li2022volunteer}. Because no one can read every message, moderators must triage within a limited reading budget. They decide \emph{where to look} (which discussions to open first), \emph{what to read} (which messages to inspect in each discussion), and \emph{how to check} (how to confirm each judgment in the original thread).

\begin{figure}[tp]
  \centering
  \includegraphics[width=0.65\linewidth]{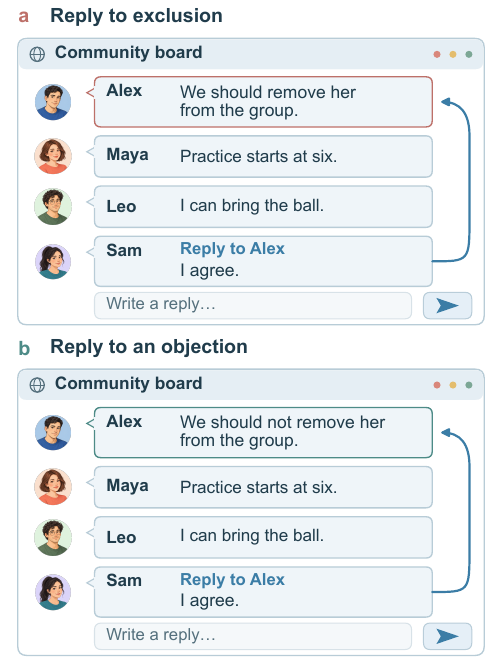}
  \caption{Reply context for moderation. ``I agree.'' supports removal in (a) and opposes it in (b).}
  \Description{Two Web discussion panels contain four messages each. Alex proposes removing a member in panel a and opposes removing her in panel b. Maya says practice starts at six; Leo offers to bring the ball. Sam then writes I agree. In each panel, a Reply to Alex label and a connecting line link Sam to Alex across the two intervening messages. Four distinct human avatars identify the speakers, consistently across panels.}
  \label{fig:motivation}
\end{figure}

These decisions are hard because meaning depends on context. The same reply, ``I agree.'', may endorse an exclusionary remark or support someone objecting to it (Figure~\ref{fig:motivation}); its wording alone does not reveal which stance it takes. A risk score can tell moderators where to look, but not what to read or how to check it. Effective triage must therefore deliver \emph{source-grounded evidence}: a few key messages that keep their original wording and reply links, so that moderators can verify them in the original discussion.

Existing work addresses these decisions separately. Risk classifiers detect personal attacks, hate speech, and other harmful content~\cite{wulczyn2017exmachina,davidson2017hate,vidgen2021dynamic}. While they tell moderators where to look, they score each message on its own and are judged by classification metrics, and adding conversational context does not consistently improve them~\cite{pavlopoulos2020context}. Budgeted selection and topic models decide what to show, either by balancing relevance against redundancy~\cite{carbonell1998mmr} or by organizing content into topics~\cite{blei2003lda,yan2013btm,grootendorst2022bertopic}, but they are designed and evaluated for coverage and topic quality~\cite{roder2015coherence,hoyle2021coherence} rather than for recovering harmful evidence. Explanation methods highlight the words behind a prediction~\cite{lei2016rationalizing,deyoung2020eraser,mathew2021hatexplain}, but they are mainly judged by whether they reflect the model's reasoning~\cite{jacovi2020faithfulness}, not by whether moderators can locate and check the evidence in the source thread. Each line improves one step, yet rarely measures whether moderators can find and verify harmful evidence within a reading budget. Chaining a classifier with a generic topic model does not close this gap: in our model-judged comparisons, replacing our grouping with BERTopic lowers all five rated dimensions, including stance preservation and evidence support (Figure~\ref{fig:judges}).

\begin{figure}[tp]
  \centering
  \includegraphics[width=0.65\linewidth]{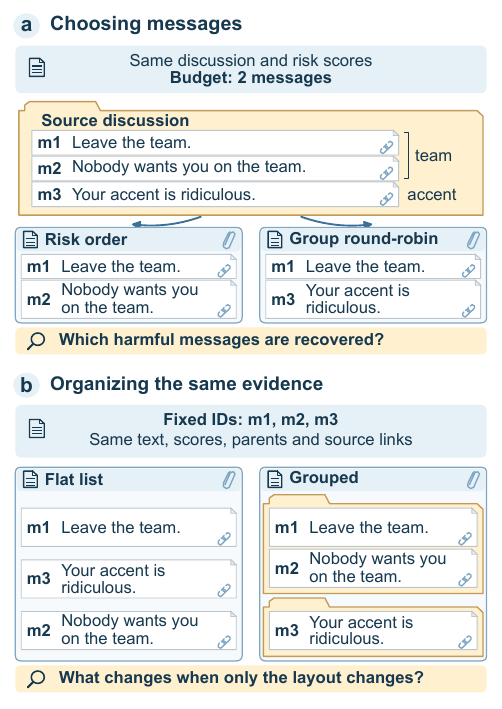}
  \caption{Two controlled comparisons. (a) Selection can change message IDs under the same discussions, risk scores, and budget. (b) A separate comparison fixes three messages and their text, scores, parents, and source links.}
  \Description{The selection panel contrasts two messages chosen by risk order with two chosen across groups. The organization panel separately fixes messages m1, m2 and m3, showing identical wording in a flat list and unnamed groups. No performance values are depicted.}
  \label{fig:controls}
\end{figure}

We observe that what to read and how to check it can be decided without changing the risk scores. Reading strictly by risk tends to spend a small budget on similar messages about the same point. In the example of Figure~\ref{fig:controls}a, the two highest-risk messages both concern the team, whereas choosing across groups also surfaces a separate attack on the target's accent. Grouping related messages lets the same budget cover distinct harms, and keeping each message tied to its source lets moderators check it in context. Because grouping leaves the scores unchanged, its benefit can be measured on its own.

Based on this observation, we present \method{}, a lightweight triage pipeline that first decides where to look and then what to show (Figure~\ref{fig:framework}). To decide where to look, \method{} pairs character patterns, which capture exact wording, with sentence meaning from a frozen multilingual encoder~\cite{reimers2020multilingual,wang2020minilm}, so that it works across languages and runs on a CPU without fine-tuning. Because each corpus labels different risk categories, \method{} learns from whichever labels a corpus provides rather than treating missing labels as negatives. To decide what to show, an independent branch groups related messages, names each group with a phrase taken from the original text, and selects messages from each group in turn. Every selected message keeps its identifier, original wording, and reply link. Unlike a generic topic model, this branch protects stance cues such as negation and counter-speech, and it never alters the risk scores.

We evaluate \method{} on five corpora drawn from Weibo discussions, online game chat, and social-media comments. SCCD~\cite{sccd2025}, a session-based Chinese cyberbullying dataset collected from Weibo, provides discussion-level labels and serves as our main benchmark for ranking discussions and selecting evidence. CONDA~\cite{conda2021}, ToxiCN~\cite{toxicn2023}, HateBR~\cite{hatebr2022}, and ScamSpot~\cite{scamspotrepo} add message-level labels for offensive language, hate speech, and spam or scams; together with SCCD, they form six message classification tasks. Beyond classification, we count how many harmful messages each selection strategy finds under the same reading budget, ask two large language models (LLMs) to judge grouped and flat layouts of the same evidence (Figure~\ref{fig:controls}b), run a user study that measures review time and decision accuracy, and time the complete pipeline on a CPU. Our contributions are:

\begin{itemize}
\item \textbf{Source-grounded triage.} We frame discussion triage as three decisions (where to look, what to read, and how to check) and present \method{}, a lightweight pipeline that ranks discussions, spreads a reading budget across groups of related messages, and links every selected message back to its source. Because grouping never alters risk scores, controlled comparisons measure ranking, selection, and presentation separately (Figure~\ref{fig:controls}).
\item \textbf{Accurate and inexpensive risk ranking across corpora.} \method{} achieves higher average precision (AP) than a fully fine-tuned XLM-R~\cite{conneau2020xlmr} on all six message tasks and exceeds a frozen-MiniLM baseline with logistic regression by 0.027--0.193 AP, while its complete pipeline runs 2.6$\times$ faster than XLM-R on a CPU. It ranks SCCD discussions with an AP of 0.976 and remains ahead of the baselines on subsets that exclude near-duplicate discussions.
\item \textbf{Better evidence and faster review.} Under the same reading budget, choosing across groups recovers more harmful messages than reading by risk, by up to 3.3 percentage points of recall, and LLM judges rate grouped layouts of the same evidence higher for clarity, stance preservation, and traceability. In a user study, reviewers needed at least 27.3\% less time with grouped evidence than with a reference display, with decision accuracy of at least 94.7\% and source-location success of at least 97.3\%. Code and dataset reconstruction resources will be released.
\end{itemize}

\section{Related Work}
\label{sec:related}

\paragraph{\normalfont\bfseries Large-model and agent safety.}
Surveys and systematic analyses map threats across large-model families and throughout data preparation, training, deployment, and agent self-evolution~\cite{ma2025safetyatscale,wang2025fullstacksafety,lin2026selfevolvingsafety}. AgentHazard evaluates harmful behavior that emerges from sequences of locally legitimate actions and accumulated context~\cite{feng2026agenthazard}. JustAsk shows how autonomous code agents can recover hidden system prompts through interaction~\cite{zheng2026justask}. These studies highlight how interaction context shapes safety risks. For Web discussions, \method{} supports human review by ranking discussions, selecting evidence under a reading budget, and preserving links to the original messages and replies.

\paragraph{\normalfont\bfseries Risk screening and context.}
Research on Web discussions covers attack measurement~\cite{wulczyn2017exmachina}, context-dependent trolling~\cite{cheng2017troll}, and early derailment prediction~\cite{zhang2018awry}. Context affects judgments of hate, counter-speech, and neutral replies~\cite{yu2022context}, although adding it does not consistently improve toxicity classifiers~\cite{pavlopoulos2020context}. Offensive language and hate speech require distinct labels~\cite{davidson2017hate}, and functional or adversarial tests expose errors hidden by aggregate scores~\cite{rottger2021hatecheck,vidgen2021dynamic}. Sentence embeddings provide reusable representations~\cite{reimers2019sbert,reimers2020multilingual}, while multilingual fine-tuning provides a supervised classification reference~\cite{conneau2020xlmr}. Unlike these classifiers, which score messages one at a time, \method{} learns from each source's own labels and uses the message scores to rank whole discussions and to select evidence that moderators can check in the original thread.

\paragraph{\normalfont\bfseries Selection under budgets and topic organization.}
Under a message budget, MMR balances relevance against redundancy in the selected material~\cite{carbonell1998mmr}. Topic methods organize content through word distributions and document--term factors, as in LDA and NMF~\cite{blei2003lda,lee1999nmf}. BTM and GSDMM target short texts~\cite{yan2013btm,yin2014gsdmm}, whereas semantic methods use embeddings, context, or joint topic discovery~\cite{dieng2020etm,bianchi2021contextual,angelov2020top2vec}. BERTopic combines sentence embeddings, clustering, and class-based term weighting~\cite{grootendorst2022bertopic}. TopicGPT, PRISM, and HILTON respectively address readable descriptions~\cite{pham2024topicgpt}, sparse semantic supervision~\cite{douglas2026prism}, and interactive analysis~\cite{chen2026hilton}. These methods are evaluated by lexical coherence~\cite{roder2015coherence} or by agreement with human judgments of topic quality~\cite{chang2009interpret,hoyle2021coherence}, not by how much harmful evidence a reader recovers within a budget. We use grouping for that purpose: to spread a fixed reading budget across distinct harms.

\paragraph{\normalfont\bfseries Evidence, explanations, and practical utility.}
By evidence, we mean the original messages and their source locations, not model rationales. \method{} retains message identifiers and character spans for direct source inspection. Learned rationalizers select text sufficient for prediction~\cite{lei2016rationalizing}, while explanation-based regularization can change a classifier's use of context~\cite{kennedy2020contextualizing}. ERASER and HateXplain evaluate rationales alongside predictions~\cite{deyoung2020eraser,mathew2021hatexplain}; Supervised Rational Attention trains attention against human rationales~\cite{eilertsen2026rationales}. Faithfulness asks whether an explanation reflects the classifier's prediction process~\cite{jain2019attention,jacovi2020faithfulness}; we ask instead whether a moderator can find the cited message and its reply context in the original thread. Human studies link explanation utility to specific tasks, including subsequent Reddit behavior~\cite{jhaver2019transparency}, professional moderators' decision speed~\cite{calabrese2024structured}, and decision quality~\cite{zhang2023biasx}. Language-model judging offers complementary assessments~\cite{liu2023geval}, with systematic biases~\cite{zheng2023judge} and sensitivity to artifacts~\cite{chen2025safer}. We therefore pair model judgments with a human study of review time and accuracy.

\section{Risk Screening and Evidence Organization}
\label{sec:method}

\begin{figure}[t]
  \centering
  \includegraphics[width=\linewidth]{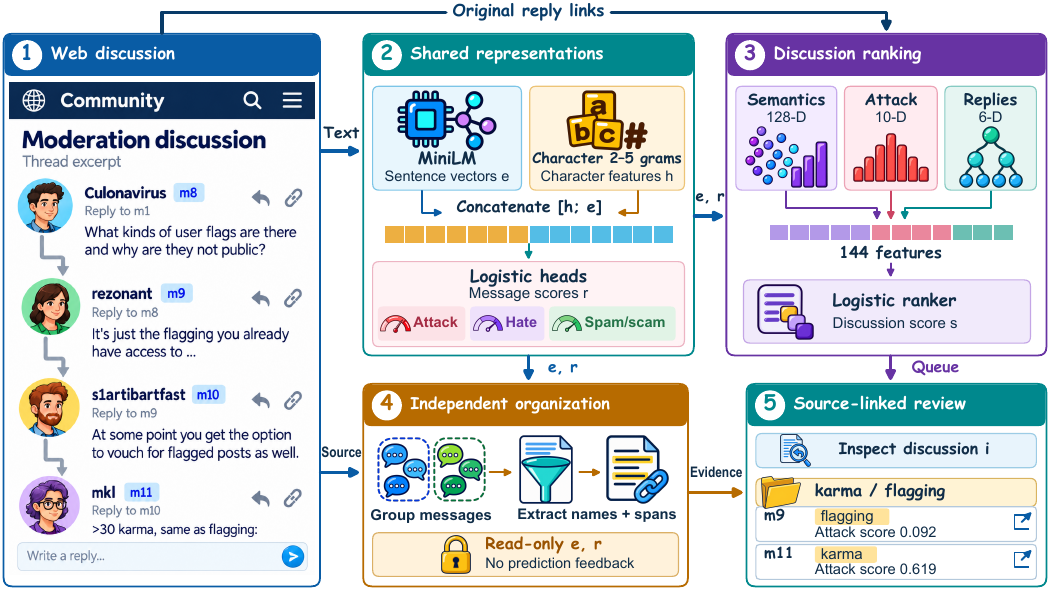}
  \caption{\method{} framework for risk screening and evidence organization. Shared character and semantic features support message risk scoring; message scores, semantic summaries, and reply relations inform discussion ranking. The organization branch groups messages and extracts phrases with source spans, enabling evidence inspection within the original reply thread.}
  \Description{A reconstructed Web discussion on the left shows the source reply chain m8 to m9 to m10 to m11, with m8 replying to m1 outside the excerpt. Illustrated avatars are accompanied by the original account handles Culonavirus, rezonant, s1artibartfast and mkl; the m labels identify messages. Distinct icons represent MiniLM sentence vectors, character features, logistic risk heads, semantic aggregation, attack statistics, reply relations, clustering and source lookup. Strong arrows route original text into shared features, original parent links into ranking, and shared sentence vectors and read-only attack scores into both ranking and organization. Source text, IDs and parents remain available for extractive organization. Colored feature strips are schematic and do not encode dimensional proportions. Ranking supplies a symbolic discussion queue; organization supplies source-linked evidence without prediction feedback. The review pane shows only m9 and m11 in the karma and flagging group, with their recorded attack scores 0.092 and 0.619 and links to the original comments.}
  \label{fig:framework}
\end{figure}

\subsection{Task and computational structure}
Discussion $C_i=\{(\mathrm{id}_{it},x_{it},p_{it},o_{it})\}_{t=1}^{n_i}$ contains each message's identifier, original text, optional direct parent, and observed order. The outputs are category-specific message scores, a discussion ranking, and source-linked evidence groups. Risk categories cover attack/offensiveness, hate/discrimination, and spam/scam promotion. Categories may overlap, whereas basic evidence groups have exclusive membership. Comments without a parent are treated as standalone messages within their original discussion.

\method{} has three computational layers (Figure~\ref{fig:framework}). Shared message representations feed partially supervised risk heads. A discussion classifier aggregates message risks, semantic features, and genuine reply relations to decide which discussions to inspect first. An independent organization branch uses the same semantic vectors to group messages, extract names, and retain source spans. This branch never updates message or discussion scores, so its contribution can be evaluated under fixed predictions and reading budgets.

\subsection{Partially supervised message screening}
A frozen multilingual MiniLM sentence encoder~\cite{reimers2019sbert,reimers2020multilingual,wang2020minilm,minilmModel} maps each message to $e_{it}\in\mathbb{R}^{384}$. Overlapping blocks cover the full text without tail truncation; their embeddings are averaged and L2-normalized. A complementary branch hashes character 2--5-grams into a vector $h_{it}$ with 65,536 nonnegative count bins~\cite{weinberger2009hashing}. The two branches are normalized separately and concatenated:
\begin{equation}
u_{it}=[h_{it};e_{it}]\in\mathbb{R}^{65,920}.
\label{eq:joint}
\end{equation}
The combined vector receives no further normalization. The sources annotate different risk categories: $m_{itc}\in\{0,1\}$ records whether label $y_{itc}$ is observed, and only $\mathcal{O}_c=\{(i,t):m_{itc}=1\}$ contributes to head $c$. Missing labels are never treated as negatives. Weights $w_{itc}$ give each contributing source equal total weight, divided equally among its observed classes. Source identifiers determine these weights but are excluded from prediction features.

Each category has a logistic head $r_{itc}=\sigma(\theta_c^{\mathsf T}u_{it}+b_c)$. For the intercept-augmented parameter $\widetilde\theta_c=[\theta_c;b_c]$, training minimizes
\begin{equation}
\frac{1}{2}\|\widetilde\theta_c\|_2^2+
C\sum_{(i,t)\in\mathcal{O}_c}w_{itc}\,
\ell(y_{itc},r_{itc}),
\label{eq:objective}
\end{equation}
where $\ell$ is binary cross-entropy. Appendix~\ref{app:implementation} specifies the block construction, weights, and LIBLINEAR~\cite{fan2008liblinear} fitting settings.

\subsection{Discussion ranking with cross-fitted supervision}
SCCD discussion labels train a separate discussion classifier and are used only at the discussion level. Ten features summarize attack scores $a_{it}=r_{it,\mathrm{attack}}$ and discussion length. A training-fitted 64-dimensional PCA projection supplies semantic vectors $v_{it}$, whose elementwise mean and maximum give 128 features. Six further features summarize semantic and risk differences across resolvable, non-self parent edges and their coverage. The resulting representation is
\begin{equation}
\begin{split}
z_i&=[z_i^{\mathrm{risk}};z_i^{\mathrm{sem}};z_i^{\mathrm{reply}}]
\in\mathbb{R}^{144},\\
z_i^{\mathrm{sem}}&=[\operatorname{mean}_t v_{it};\max_t v_{it}].
\end{split}
\label{eq:discussion}
\end{equation}
A class-balanced logistic head maps the training-standardized vector to score $s_i$. Appendix~\ref{app:implementation} defines the summary statistics, including defaults for discussions with no valid parent edges.

Training uses message scores from three-fold cross-fitting over SCCD discussions. Each risk-head fit excludes labels from its held-out discussions and matching duplicate components across all sources. Only supervised risk heads are cross-fitted; the same encoder and unsupervised, training-fitted PCA are shared across folds. Out-of-fold scores train the discussion classifier.

At deployment, only the message heads are refitted, on the full eligible training partition; they score new discussions for the standardizer and discussion classifier fitted during training. Thresholds maximize validation macro-F1, and rankings use the continuous scores directly.

\subsection{Extractive discussion groups}
Within each discussion, the organization branch extracts contiguous phrases with original character spans. It excludes generic words and content-free acknowledgements, and protects negated or interrogative clauses from misleading fragments. Messages with nonzero embeddings and usable phrases enter complete-link clustering with cosine distance and a merge threshold of 0.35. A fixed lexical rule sets the distance to one between some counter-speech and high-risk attacks. The rule uses lexical patterns for Chinese, English, and Portuguese.

Groups require at least two basic members and a usable extractive name. Candidate names are ranked by within-group support, cross-group specificity, and capped phrase length. They retain message IDs and Unicode spans. A short reply without a usable phrase can join only its direct parent's group; it keeps its own risk score, does not affect naming, and passes no association on to its own replies. Ungrouped messages remain selectable. Appendix~\ref{app:implementation} gives the complete eligibility and naming rules.

Figure~\ref{fig:framework} illustrates this output with excerpts from a saved Hacker News snapshot: \emph{karma / flagging} links $m_9$ and $m_{11}$ to their original spans while retaining the intervening reply chain. The example is illustrative and not part of the benchmark; Appendix~\ref{app:web-illustration} records its provenance and inference parameters.

\subsection{Evidence selection under reading budgets}
The screening protocol ranks discussions by $s_i$ and selects at most $B$ messages from each of the first $L$ discussions. Four selectors share these discussion ranks and attack scores: original order, descending risk, maximal marginal relevance (MMR), and group round-robin. MMR~\cite{carbonell1998mmr} balances risk against similarity to already selected messages. Group round-robin orders groups by their highest-risk member, then cycles through their internal risk order; ungrouped messages form singleton queues. Appendix~\ref{app:budget-accounting} specifies MMR, tie-breaking, and character accounting.

The message budget counts selected messages only. A complementary character budget also counts displayed direct parents, titles, and group headings.

For the selected messages $\mathcal S_{L,B}$ with observed attack labels $y_{it}$, message-level precision and recall are defined as
\begin{equation}
\begin{split}
P_{L,B}&=\frac{\sum_{(i,t)\in\mathcal S_{L,B}}y_{it}}{|\mathcal S_{L,B}|},\\
R_{L,B}&=\frac{\sum_{(i,t)\in\mathcal S_{L,B}}y_{it}}
 {\sum_{(i,t)\in\mathcal T}y_{it}},
\end{split}
\label{eq:budget-endpoints}
\end{equation}
where $\mathcal T$ contains all labeled test messages. All SCCD test messages have observed attack labels, so every selected message enters the precision denominator. Precision measures the concentration of labeled dangerous messages in the selected set; recall measures recovery from the entire test corpus, including unopened discussions. Displayed parents add reading cost but count toward selected evidence only when selected themselves.

\section{Experimental Protocol}
\label{sec:protocol}
\begin{table}[t]
\centering
\normalsize
\caption{Message counts and observed test labels. A/H/S denote attack/offense, hate, and spam/scam. Positive rates use observed labels; ToxiCN counts apply to each task.}
\label{tab:datasets}
\setlength{\tabcolsep}{3.4pt}
\renewcommand{\arraystretch}{1.0}
\begin{tabularx}{\linewidth}{Xrrrcrr}
\toprule
\rowcolor{CSTTableHeader}
& \multicolumn{3}{c}{\textbf{Split sizes}} & \multicolumn{3}{c}{\textbf{Observed test labels}} \\
\rowcolor{CSTTableHeader}
\multirow{-2}{*}{\textbf{Source}} & \textbf{Train} & \textbf{Dev.} & \textbf{Test} & \textbf{Task} & \textbf{Count} & \textbf{Pos. (\%)} \\
\midrule
\cstSource{sccd}{SCCD} & 21,762 & 7,849 & 8,028 & A & 8,028 & 24.4 \\
\rowcolor{CSTTableStripe}
\cstSource{conda}{CONDA} & 14,547 & 4,315 & 4,746 & A & 4,746 & 23.8 \\
\cstSource{toxicn}{ToxiCN} & 7,720 & 1,866 & 2,409 & A/H & 2,212 & 48.6/41.2 \\
\rowcolor{CSTTableStripe}
\cstSource{scamspot}{ScamSpot} & 1,847 & 623 & 582 & S & 582 & 32.0 \\
\cstSource{hatebr}{HateBR} & 4,129 & 1,154 & 1,717 & A & 1,717 & 59.6 \\
\bottomrule
\end{tabularx}
\end{table}

\paragraph{\normalfont\bfseries Data and splits.}
The cleaned corpus contains 83,294 messages from five sources~\cite{sccd2025,conda2021,toxicn2023,hatebr2022,scamspotrepo} (Table~\ref{tab:datasets}). SCCD provides Weibo discussions, CONDA provides Dota~2 game chat, and ToxiCN, HateBR, and ScamSpot provide social-media comments. ScamSpot combines spam and scam labels. Unknown labels are excluded separately for each risk category. These corpora support message classification, with SCCD providing the primary discussion benchmark.

We preserve SCCD's 406/135/136 train/validation/test discussion split and ToxiCN's official test boundary. CONDA matches and HateBR posts define sampling units; duplicate components constrain the remaining splits. Cross-partition text matches are removed from training, and conflicting training supervision is masked while original evaluation labels are retained. Normalized exact matches and character-5-gram near matches define duplicate components. Evaluation clusters merge components connected through sampling units, including cross-source links. The SCCD test set contains 47 clusters, with the largest covering 89 of 136 discussions. Full-cohort comparisons therefore include isolated-subset and largest-component-removed evaluations. Because these fixed partitions were examined in earlier risk evaluations, we confirm the main results on a separately acquired cohort.

\paragraph{\normalfont\bfseries Classification baselines and training.}
LDA, BTM, GSDMM, NMF, and BERTopic provide topic representations with supervised logistic-regression output heads. They share the joint training data, observed-label masks, and weighting rule. Traditional methods use 32 topics; BERTopic determines its topic count from training data. MiniLM-plus-LR shares the semantic encoder and supervision with \method{} and omits character features. Discussion comparisons aggregate each method's own representation. Fully fine-tuned XLM-R-base~\cite{conneau2020xlmr} participates in message classification and complete-service CPU comparisons under the same observed-label supervision.

Average precision (AP) is the primary metric for message and discussion ranking under class imbalance~\cite{saito2015pr}. Training uses fixed search budgets~\cite{dodge2019reporting} and seeds 42, 43, and 44. Training partitions fit representations, validation selects operating thresholds, and mean validation AP across six tasks selects XLM-R checkpoints. AP evaluates the full ranking; thresholded metrics describe the validation-selected operating point. Four leave-one-source-out attack tasks assess message transfer by holding out SCCD, CONDA, ToxiCN, or HateBR in turn. Joint, character-only, and semantic-only representations use the same held-out source and evaluation labels within each comparison. Appendix~\ref{app:implementation} gives the fitting settings and metric definitions.

\paragraph{\normalfont\bfseries Reading budgets and evidence metrics.}
Original order, risk order, MMR, and group round-robin share discussion membership, risk scores, and reading budgets within each comparison. The screening grid uses $L\in\{10,20,30\}$ discussions and $B\in\{1,3,5,10\}$ messages per discussion. The primary contrast compares group round-robin with risk order at $L=20,B=3$; original-order and MMR comparisons are secondary. Evidence hits use the original message-level attack labels. Global recall divides selected dangerous messages by all 1,960 positive test messages; evidence precision divides them by the actual number of selected messages. Message- and discussion-level outcomes differ: 101 test discussions contain at least one positive message, but only 71 carry a positive discussion label.

The evidence-only comparison applies $B$ to every discussion without a top-$L$ cutoff. Its full-cohort and isolated-subset recall denominators are 1,960 and 413 positive messages, respectively. Selected-message budgets and displayed-text costs are recorded separately. Character accounting includes displayed context under the cost rules in Section~\ref{sec:method}.

\paragraph{\normalfont\bfseries Model assessment and human review.}
The model evaluation includes 40 independent SCCD discussions and seven repeated cases per judge (Table~\ref{tab:judge-coverage}). Cases exclude dependencies with training and development material. Gemini~2.5~Flash and GPT-4o-mini first label raw discussions without algorithm outputs, then assess anonymous configurations in fresh contexts. The seven configurations comprise \method{}, direct semantic clustering, BERTopic grouping, frequency naming, and separate removals of parent association, polarity protection, or counter-speech separation.

Judges score naming, group consistency, coverage, stance preservation, and source support on $0/1/2$, with unknown available for abstention. Non-abstaining judgments require locally validated evidence identifiers. Method names and the other judge's responses are hidden; balanced presentation addresses order sensitivity~\cite{wang2024fair}. The fixed-evidence comparison holds messages, scores, parent text, and displayed wording constant, varying only order and unnamed group boundaries. Comparisons use jointly scoreable cases, configurations, and dimensions, with cross-model analyses restricted to the same cases and unknown judgments excluded.

The human study compares grouped and reference displays under equal reading budgets. Reviewers receive different, difficulty-matched discussions with counterbalanced condition order and no repeated exposure to the same discussion. Outcomes are completion time, risk-decision accuracy, missed-risk rate, and source-location success, with separate denominators for decisions and location queries. A separate semantic audit covers 36 independent discussions across three platforms, with 12 discussions per platform and at least two assessors. It evaluates name meaning, stance preservation, and evidence support alongside checks of message identifiers, source spans, and parent links. Appendix~\ref{app:judge-rubric} specifies the rubrics, paired-score denominators, human tasks, and resampling design.

\paragraph{\normalfont\bfseries Further validation.}
Discussion-level evaluation includes at least two additional sources, each with at least 200 independent discussions. \method{} and MiniLM share source labels, sampling units, and test discussions within each source. Ten dependency-cluster repartitions of development material assess stability while preserving official test boundaries. Each repartition fits representations on its training partition and selects settings on its validation partition. Final confirmation fixes model settings, thresholds, comparison families, sampling rules, and analysis settings before opening a separately acquired cohort. Table~\ref{tab:extended-evaluation} summarizes these evaluations, and Appendix~\ref{app:extended-evaluation} details their scope.

\paragraph{\normalfont\bfseries Statistical analysis.}
Paired bootstrap resampling~\cite{efron1993bootstrap} follows the sampling design~\cite{dror2018testing}, using fixed dependency clusters as the resampling units. Each screening resample reranks discussions before applying $L$. The budget family contains three contrasts against risk order, three values of $L$, and four values of $B$, yielding 36 comparisons per metric. Six-task comparisons and the eight transfer contrasts use their respective adjustment families. Seed-level variation and paired cluster intervals quantify training variability and test-sample uncertainty, respectively; development repartitions share observations and are summarized separately.

\paragraph{\normalfont\bfseries CPU measurement.}
MiniLM, \method{}, and XLM-R run on the same four-thread hardware over 136 SCCD discussions containing 8,028 messages, with three passes per method. The matched service includes raw-text encoding, risk scoring, discussion processing, grouping, evidence selection, and serialization. Batch timings exclude asset loading, warm-up, and disk writes. Separate startup measurements report loading and first-request latency under matched cache conditions. Appendix~\ref{app:implementation} specifies the hardware, process-memory accounting, and startup conditions.

\section{Results}
\label{sec:results}

\subsection{Evidence recovery under matched reading budgets}
\begin{table}[!htb]
\centering
\normalsize
\caption{Primary screening at $L=20,B=3$, selecting at most 60 messages from the same 20 discussions. Recall covers all 1,960 dangerous test messages.}
\label{tab:budget-main}
\setlength{\tabcolsep}{5pt}
\renewcommand{\arraystretch}{1.0}
\begin{tabularx}{\linewidth}{Xrr}
\toprule
\rowcolor{CSTTableHeader}
\textbf{Selector} & \shortstack[r]{\textbf{Evidence}\\\textbf{precision (\%)}} & \shortstack[r]{\textbf{Global}\\\textbf{recall (\%)}} \\
\midrule
\rowcolor{CSTTableStripe}
\cstSelector{original}{Original order} & 47.778 & 1.463 \\
\cstSelector{risk}{Risk order} & 91.667 & 2.806 \\
\rowcolor{CSTTableStripe}
\cstSelector{mmr}{MMR} & 90.000 & 2.755 \\
\cstSelector{group}{Group round-robin} & 96.667 & 2.959 \\
\bottomrule
\end{tabularx}
\par\smallskip
\begin{tabularx}{\linewidth}{Xr}
\toprule
\rowcolor{CSTTableFocus}
\textbf{Designated contrast: group round-robin $-$ risk} & \textbf{$\Delta$ recall (pp)} \\
\midrule
Same 20 discussions and 60 selected messages & \cstDeltaPos{+0.153} \\
\bottomrule
\end{tabularx}
\par\smallskip\raggedright pp: percentage points. Group round-robin and risk order recover 58 and 55 dangerous messages on average, respectively, from a denominator of 1,960.
\end{table}

With the same discussion queue, risk scores, and reading budget, group round-robin recovers more dangerous messages than risk order (Table~\ref{tab:budget-main}). At the primary setting $L=20,B=3$, both select at most 60 messages; group round-robin finds an average of \keyresult{58 dangerous messages versus 55}, and selected-evidence precision rises from 91.667\% to 96.667\%. Global recall rises from 2.806\% to \keyresult{2.959\%} (\keyresult{+0.153 percentage points}); it is low for every selector because 60 messages can cover at most 3.061\% of the 1,960 dangerous test messages. Group round-robin also exceeds original order and MMR at this setting.

On the independent confirmation cohort, the adjusted paired intervals for the designated recall gains stay above zero (Table~\ref{tab:extended-evaluation}).

\begin{figure}[t]
\centering
\includegraphics[width=\textwidth]{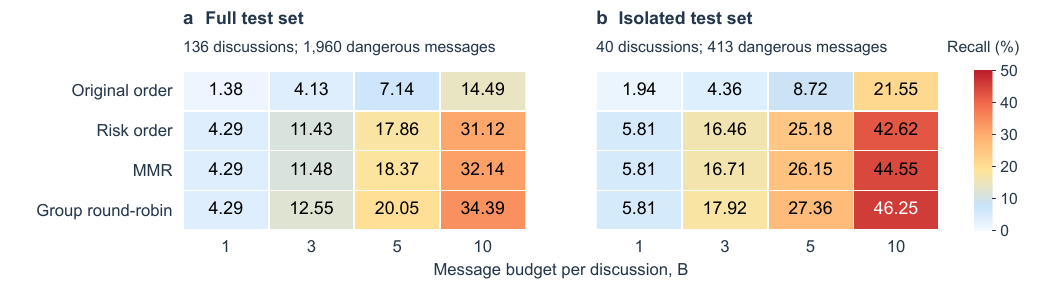}
\caption{Evidence-only budget sensitivity. Cells show recall (\%) across message budgets $B$, using 1,960 dangerous messages in the full test set (a) and 413 in the isolated set (b) as denominators. Every discussion receives a budget; this differs from the top-$L$ screening endpoint.}
\Description{Two annotated recall heatmaps compare original order, risk order, MMR, and group round-robin at budgets of one, three, five, and ten messages per discussion. Risk-based selectors coincide at budget one. Group round-robin has the highest recall at larger budgets in both cohorts, reaching 34.39 percent on the full cohort and 46.25 percent on the isolated cohort at budget ten.}
\label{fig:budget}
\end{figure}

Figure~\ref{fig:budget} evaluates selection across every discussion, extending the comparison beyond top-$L$ screening. Group round-robin has the highest recall at $B=3,5,10$ in both the full and isolated cohorts. Its gains over risk order range from 1.122 to 3.265 percentage points in the full cohort and from 1.453 to 3.632 points in the isolated cohort. At $B=10$, recall reaches \keyresult{34.39\%} and \keyresult{46.25\%}, respectively. All three risk-based selectors coincide at $B=1$, when they select the highest-risk message. The gap widens as the budget grows and selections reach more groups.

\subsection{Joint features improve message AP}
\begin{table}[t]
\centering
\normalsize
\definecolor{CSTTableTopic}{HTML}{DADDE3}
\definecolor{CSTTableFeatures}{HTML}{CFEAF6}
\definecolor{CSTTableTuned}{HTML}{FBD9C9}
\caption{Message test AP. Topic representations are evaluated with supervised LR adapters.}
\label{tab:message-ap}
\setlength{\tabcolsep}{4pt}
\begin{tabularx}{.9\textwidth}{l*{6}{>{\raggedleft\arraybackslash}X}}
\toprule
Method & SCCD & CONDA & HateBR & ToxiCN-A & ToxiCN-H & ScamSpot \\
\midrule
\rowcolor{CSTTableTopic!25}
LDA & 0.307 & 0.363 & 0.642 & 0.621 & 0.623 & 0.829 \\
\rowcolor{CSTTableTopic!25}
BTM & 0.356 & 0.300 & 0.842 & 0.589 & 0.534 & 0.731 \\
\rowcolor{CSTTableTopic!25}
GSDMM & 0.402 & 0.385 & 0.643 & 0.597 & 0.567 & 0.630 \\
\rowcolor{CSTTableTopic!25}
NMF & 0.329 & 0.609 & 0.729 & 0.541 & 0.487 & 0.606 \\
\rowcolor{CSTTableTopic!25}
BERTopic & 0.354 & 0.413 & 0.698 & 0.592 & 0.542 & 0.699 \\
\rowcolor{CSTTableFeatures!45}
MiniLM--LR & 0.597 & 0.694 & 0.951 & 0.758 & 0.735 & 0.954 \\
\rowcolor{CSTTableFeatures!45}
XLM-R & 0.683 & 0.859 & 0.978 & 0.864 & 0.854 & 0.972 \\
\rowcolor{CSTTableTuned!45}
\textbf{\method{}} & 0.712 & 0.887 & 0.986 & 0.893 & 0.879 & 0.981 \\
\bottomrule
\end{tabularx}
\end{table}

\method{} exceeds MiniLM-plus-LR on all six message-classification tasks by \keyresult{0.027--0.193 absolute AP}, with the largest gain on CONDA (Table~\ref{tab:message-ap}). It also outperforms the five topic representations under their common supervised adapter. Relative to fine-tuned XLM-R, AP gains range from 0.008 to 0.029; the smallest gains occur on HateBR and ScamSpot. These gains require no fine-tuning of the sentence encoder.

The SCCD ablations show that character and semantic features contribute complementary predictive information (Table~\ref{tab:ablations}). Removing characters lowers message AP from 0.712 to 0.597, and removing message semantics lowers it to 0.524. Source balancing contributes a further 0.026 AP relative to its removal. At the validation-selected operating threshold, \method{} achieves precision 0.719, recall 0.713, and macro-F1 0.812 on SCCD (Table~\ref{tab:extended-evaluation}).

The results are stable across ten dependency-cluster repartitions of development material: taskwise AP sample SDs are at most 0.009, with at least nine paired wins over MiniLM. On the independent confirmation cohort, the adjusted paired intervals for the designated AP gains stay above zero (Table~\ref{tab:extended-evaluation}).

\subsection{Joint features support cross-source transfer}
Without target-source supervision, leave-one-source-out attack AP reaches 0.657 on SCCD, 0.633 on CONDA, 0.852 on ToxiCN, and 0.954 on HateBR (Figure~\ref{fig:stability-transfer}). The joint representation exceeds MiniLM by \keyresult{0.098, 0.073, 0.141, and 0.025 AP}, respectively, and also exceeds the character-only reference on every source. Relative to in-domain attack classification, AP falls by 0.055, 0.254, 0.041, and 0.032; CONDA shows the largest drop.

\subsection{Discussion ranking on SCCD and external sources}
\begin{table}[t]
\centering
\normalsize
\caption{Discussion AP. $n$: discussions; clusters: duplicate-connected units. Development AP selects BTM as the topic baseline for comparison.}
\label{tab:conversation}
\setlength{\tabcolsep}{3pt}
\begin{tabular*}{\linewidth}{@{\extracolsep{\fill}}lrrrrr@{}}
\toprule
Cohort & $n$ & Clusters & BTM & \mbox{MiniLM--LR} & \method{} \\
\midrule
Full & 136 & 47 & 0.890 & 0.957 & 0.976 \\
Isolated & 40 & 40 & 0.819 & 0.938 & 0.963 \\
Largest removed & 47 & 46 & 0.769 & 0.912 & 0.948 \\
\bottomrule
\end{tabular*}
\end{table}

\method{} improves discussion ranking across all three SCCD cohorts (Table~\ref{tab:conversation}). Its AP is \keyresult{0.976 versus 0.957} for MiniLM-plus-LR on the full cohort, \keyresult{0.963 versus 0.938} on the 40 isolated discussions, and \keyresult{0.948 versus 0.912} after removing the largest component. The corresponding gains are 0.019, 0.025, and 0.036 AP. BTM trails by a larger margin in every cohort. The advantage therefore persists outside the largest dependency component, which contains 89 of 136 discussions and 5,855 of 8,028 test messages.

Discussion-level features contribute beyond the message scorer (Table~\ref{tab:ablations}). Removing discussion semantics, reply statistics, or risk statistics lowers discussion AP by 0.014, 0.017, or 0.033, respectively. Removing risk statistics causes the largest drop, so aggregated message risk contributes most to discussion ranking.

External evaluation extends this result to at least two additional sources, each with at least 200 independent discussions (Table~\ref{tab:extended-evaluation}). \method{} and MiniLM use the same source labels, sampling units, and test discussions in each comparison. On each source, \method{} achieves AP of at least 0.913 and exceeds MiniLM by at least 0.023 AP. These matched comparisons show that the discussion-ranking advantage extends beyond SCCD, complementing the message-level transfer findings with evidence for ranking whole discussions.

\subsection{Model judgments of evidence organization}
\begin{figure}[t]
\centering
\includegraphics[width=\columnwidth]{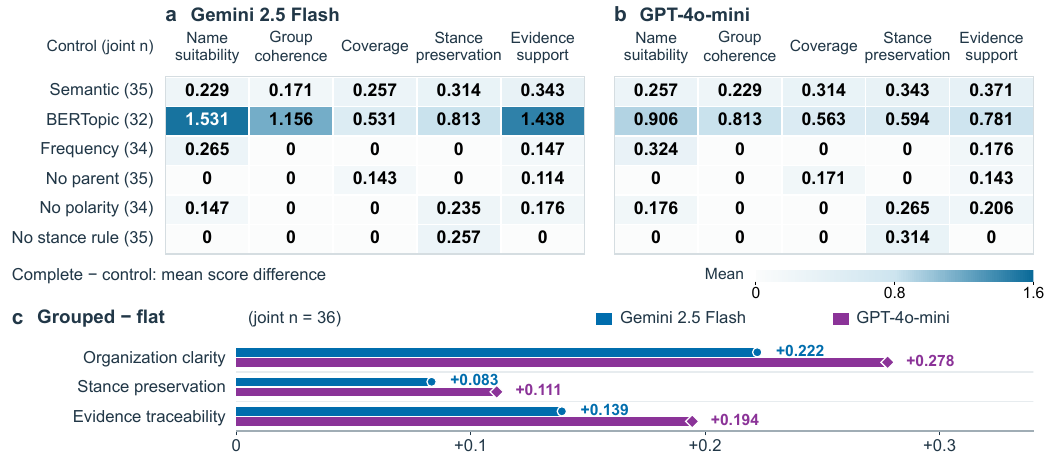}
\caption{Paired score differences on a 0--2 rubric: complete minus control (a,b) and grouped minus flat with fixed evidence (c, $n=36$).}
\Description{Two heatmaps compare the complete module with six controls for Gemini 2.5 Flash and GPT-4o-mini. The complete module has positive differences against simple semantic grouping and BERTopic. Naming, parent association, polarity protection, and counter-speech separation contribute in their corresponding dimensions, with zeros retained elsewhere. A lower panel compares grouped and flat presentations on clarity, stance preservation, and traceability, with positive means for both judges and no confidence intervals displayed.}
\label{fig:judges}
\end{figure}

Both judges favor the complete module over simple semantic grouping and BERTopic on all five dimensions (Figure~\ref{fig:judges}). On jointly scoreable cases, complete-minus-semantic-grouping differences range from 0.171 to 0.343 for Gemini and from 0.229 to 0.371 for GPT-4o-mini. The component comparisons below show where these gains come from.

Extractive naming improves name suitability over frequency naming by 0.265 for Gemini and 0.324 for GPT-4o-mini. Parent association improves coverage and source support for both judges. Polarity protection improves stance preservation by 0.235 for Gemini and 0.265 for GPT-4o-mini; counter-speech separation contributes 0.257 and 0.314, respectively.

With message identities, wording, scores, and parent context held fixed, both judges find the same evidence clearer and easier to trace when it is grouped (panel~c). Grouping improves clarity by \keyresult{0.222} for Gemini and \keyresult{0.278} for GPT-4o-mini. Stance-preservation gains are 0.083 and 0.111, and traceability gains are 0.139 and 0.194.

\subsection{Human review and source verification}
Under equal reading budgets, grouped presentation reduces review time by \keyresult{at least 27.3\%} relative to the reference display (median within-reviewer completion-time ratio $\leq 0.727$; Table~\ref{tab:extended-evaluation}). Decision accuracy is \keyresult{at least 94.7\%}, the missed-risk rate is at most 5.3\%, and reviewers locate the source in at least 97.3\% of queries.

A separate audit traces group names and cited evidence back to the original discussion in 36 independent discussions across three platforms. Every tested source mapping is correct, and at least 33 cases pass human semantic verification. Appendix~\ref{app:judge-rubric} specifies the separate study designs and denominators.

\subsection{CPU cost of the complete service}
\begin{figure}[t]
\centering
\includegraphics[width=0.72\linewidth]{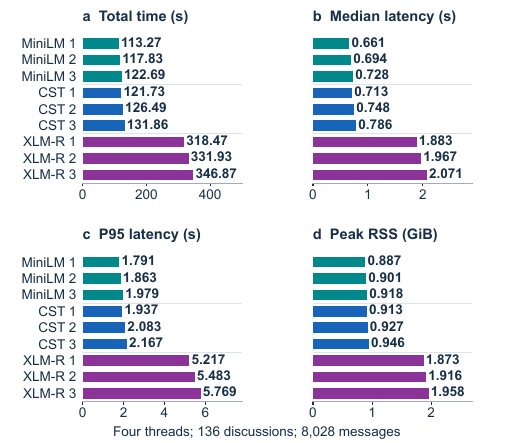}
\caption{CPU service cost on SCCD: four threads and three passes per method.}
\Description{Four horizontal-bar panels compare MiniLM, CST, and XLM-R over three passes each. CST totals are 121.73, 126.49, and 131.86 seconds, with median latencies 0.713, 0.748, and 0.786 seconds, P95 latencies 1.937, 2.083, and 2.167 seconds, and peak RSS 0.913, 0.927, and 0.946 GiB. Mean total times are 117.93, 126.69, and 332.42 seconds for MiniLM, CST, and XLM-R. P95 is a latency percentile, not a confidence interval.}
\label{fig:cpu-cost}
\end{figure}

The complete SCCD service takes \keyresult{7.43\%} more processing time than MiniLM-plus-LR and runs \keyresult{2.624$\times$} as fast as the matched XLM-R service (Figure~\ref{fig:cpu-cost}). Adding character features to the frozen semantic representation therefore costs 7.43\% more processing time and adds 0.027--0.193 AP across the six message tasks.

Across three four-thread CPU passes over 136 discussions and 8,028 messages, mean total time is 126.69 seconds for \method{}, 117.93 seconds for MiniLM-plus-LR, and 332.42 seconds for XLM-R. \method{}'s median discussion latency ranges from 0.713 to 0.786 seconds across passes, and its maximum process-tree RSS is \keyresult{0.946\,GiB}. Appendix~\ref{app:implementation} details the hardware and service scope.

Separate startup measurements give encoder loading times of at most 9.713 seconds and first-request latency of at most 1.837 seconds under matched hardware and cache conditions (Table~\ref{tab:extended-evaluation}).

\section{Discussion and Limitations}
Our results suggest that a risk score works best as a queue rather than as a reading order. With the queue and scores fixed, spreading the reading budget across groups recovers more harmful messages, and grouped, source-linked evidence lets reviewers finish faster while keeping decision accuracy high. The stance safeguards matter here: removing polarity protection or counter-speech separation lowers judged stance preservation. Character features add 0.027--0.193 message AP for 7.43\% more CPU time, so these gains do not require fine-tuning a large encoder. Separating scoring and presentation makes selectors directly comparable under fixed predictions, with shared message identifiers and source spans supporting output checks.

\paragraph{\normalfont\bfseries Limitations.}
First, the evidence-selection and model-judging results come mainly from SCCD, a single Chinese platform whose largest dependency cluster covers 89 of 136 test discussions. Second, source and language change together in the transfer tests, so their effects cannot be separated. Third, naming and grouping quality are rated mainly by two LLM judges, with a human audit of 36 discussions, and the stance rule covers only Chinese, English, and Portuguese. Finally, the user study uses controlled, budget-matched tasks rather than live moderation, where workload and incentives differ.

\paragraph{\normalfont\bfseries Ethics and availability.}
The corpora contain offensive public interactions. Scores support screening, with no sensitive-identity inference or automatic sanctions. Dialect and identity-term biases require dedicated audits~\cite{sap2019racial,dixon2018bias}. Screening runs locally; auxiliary model assessment follows provider and data-access conditions. Code and reconstruction instructions preserve original licensing and access restrictions without redistributing restricted text.

\section{Conclusion}
Risk scores tell moderators where to look, but not what to read or how to check
it. CST treats these as separate decisions: it scores messages and ranks
discussions on an ordinary CPU, then selects and groups evidence without changing
the scores, keeping every selected message linked to its source. CST achieves
higher message AP than a fully fine-tuned XLM-R on all six tasks while running
2.6$\times$ faster on a CPU. Under the same reading budget, choosing across groups
recovers up to 3.3 percentage points more harmful messages than reading by risk,
and two LLM judges rate grouped layouts of the same evidence as clearer, more
stance-preserving, and easier to trace. In a user study, reviewers needed at
least 27.3\% less time with grouped evidence while keeping decision accuracy
above 94\%. This gives moderators a continuous path from a prioritized discussion through selected evidence to the original messages and their surrounding reply context for verification. More broadly, deciding what to show can be optimized and evaluated
separately from how risk is scored.

\FloatBarrier
\label{main:end}
\bibliographystyle{plainnat}
\bibliography{references}

\begin{thebibliography}{62}
\providecommand{\natexlab}[1]{#1}
\providecommand{\url}[1]{\texttt{#1}}
\expandafter\ifx\csname urlstyle\endcsname\relax
  \providecommand{\doi}[1]{doi: #1}\else
  \providecommand{\doi}{doi: \begingroup \urlstyle{rm}\Url}\fi

\bibitem[Angelov(2020)]{angelov2020top2vec}
Dimo Angelov.
\newblock {Top2Vec}: Distributed representations of topics.
\newblock arXiv:2008.09470, 2020.
\newblock URL \url{https://arxiv.org/abs/2008.09470}.

\bibitem[Bianchi et~al.(2021)Bianchi, Terragni, and
  Hovy]{bianchi2021contextual}
Federico Bianchi, Silvia Terragni, and Dirk Hovy.
\newblock Pre-training is a hot topic: Contextualized document embeddings
  improve topic coherence.
\newblock In \emph{Proceedings of the 59th Annual Meeting of the Association
  for Computational Linguistics and the 11th International Joint Conference on
  Natural Language Processing (Volume 2: Short Papers)}, pages 759--766.
  Association for Computational Linguistics, 2021.
\newblock \doi{10.18653/v1/2021.acl-short.96}.
\newblock URL \url{https://aclanthology.org/2021.acl-short.96/}.

\bibitem[Blei et~al.(2003)Blei, Ng, and Jordan]{blei2003lda}
David~M. Blei, Andrew~Y. Ng, and Michael~I. Jordan.
\newblock Latent {Dirichlet} allocation.
\newblock \emph{Journal of Machine Learning Research}, 3:\penalty0 993--1022,
  2003.
\newblock URL \url{https://jmlr.org/papers/v3/blei03a.html}.

\bibitem[Calabrese et~al.(2024)Calabrese, Neves, Shah, Bos, Ross, Lapata, and
  Barbieri]{calabrese2024structured}
Agostina Calabrese, Leonardo Neves, Neil Shah, Maarten~W. Bos, Bj{\"o}rn Ross,
  Mirella Lapata, and Francesco Barbieri.
\newblock Explainability and hate speech: Structured explanations make social
  media moderators faster.
\newblock In \emph{Proceedings of the 62nd Annual Meeting of the {A}ssociation
  for {C}omputational {L}inguistics (Volume 2: Short Papers)}, pages 398--408.
  Association for Computational Linguistics, 2024.
\newblock \doi{10.18653/v1/2024.acl-short.38}.
\newblock URL \url{https://aclanthology.org/2024.acl-short.38/}.

\bibitem[Carbonell and Goldstein(1998)]{carbonell1998mmr}
Jaime Carbonell and Jade Goldstein.
\newblock The use of {MMR}, diversity-based reranking for reordering documents
  and producing summaries.
\newblock In \emph{Proceedings of the 21st Annual International ACM SIGIR
  Conference on Research and Development in Information Retrieval}, pages
  335--336. ACM, 1998.
\newblock \doi{10.1145/290941.291025}.

\bibitem[Chang et~al.(2009)Chang, Boyd-Graber, Gerrish, Wang, and
  Blei]{chang2009interpret}
Jonathan Chang, Jordan Boyd-Graber, Sean Gerrish, Chong Wang, and David~M.
  Blei.
\newblock Reading tea leaves: How humans interpret topic models.
\newblock In \emph{Advances in Neural Information Processing Systems},
  volume~22. Curran Associates, Inc., 2009.
\newblock URL
  \url{https://proceedings.neurips.cc/paper_files/paper/2009/file/f92586a25bb3145facd64ab20fd554ff-Paper.pdf}.

\bibitem[Chen and Goldfarb-Tarrant(2025)]{chen2025safer}
Hongyu Chen and Seraphina Goldfarb-Tarrant.
\newblock Safer or luckier? {LLM}s as safety evaluators are not robust to
  artifacts.
\newblock In \emph{Proceedings of the 63rd Annual Meeting of the {A}ssociation
  for {C}omputational {L}inguistics (Volume 1: Long Papers)}, pages
  19750--19766. Association for Computational Linguistics, 2025.
\newblock \doi{10.18653/v1/2025.acl-long.970}.
\newblock URL \url{https://aclanthology.org/2025.acl-long.970/}.

\bibitem[Chen et~al.(2026)Chen, Archasantisuk, and Bjerva]{chen2026hilton}
Yiyi Chen, Sukhumarn Archasantisuk, and Johannes Bjerva.
\newblock Towards understanding professional {AI} assistant use:
  Human-in-the-loop topic modeling of multi-turn conversations.
\newblock In \emph{Companion Proceedings of the ACM Web Conference 2026}, pages
  379--390. ACM, 2026.
\newblock \doi{10.1145/3774905.3794645}.

\bibitem[Cheng et~al.(2017)Cheng, Bernstein, Danescu-Niculescu-Mizil, and
  Leskovec]{cheng2017troll}
Justin Cheng, Michael Bernstein, Cristian Danescu-Niculescu-Mizil, and Jure
  Leskovec.
\newblock Anyone can become a troll: Causes of trolling behavior in online
  discussions.
\newblock In \emph{Proceedings of the 2017 ACM Conference on Computer Supported
  Cooperative Work and Social Computing}, pages 1217--1230. ACM, 2017.
\newblock \doi{10.1145/2998181.2998213}.
\newblock URL \url{https://doi.org/10.1145/2998181.2998213}.

\bibitem[Conneau et~al.(2020)Conneau, Khandelwal, Goyal, Chaudhary, Wenzek,
  Guzm{\'a}n, Grave, Ott, Zettlemoyer, and Stoyanov]{conneau2020xlmr}
Alexis Conneau, Kartikay Khandelwal, Naman Goyal, Vishrav Chaudhary, Guillaume
  Wenzek, Francisco Guzm{\'a}n, Edouard Grave, Myle Ott, Luke Zettlemoyer, and
  Veselin Stoyanov.
\newblock Unsupervised cross-lingual representation learning at scale.
\newblock In \emph{Proceedings of the 58th Annual Meeting of the {A}ssociation
  for {C}omputational {L}inguistics}, pages 8440--8451. Association for
  Computational Linguistics, 2020.
\newblock \doi{10.18653/v1/2020.acl-main.747}.
\newblock URL \url{https://aclanthology.org/2020.acl-main.747/}.

\bibitem[Davidson et~al.(2017)Davidson, Warmsley, Macy, and
  Weber]{davidson2017hate}
Thomas Davidson, Dana Warmsley, Michael Macy, and Ingmar Weber.
\newblock Automated hate speech detection and the problem of offensive
  language.
\newblock \emph{Proceedings of the International AAAI Conference on Web and
  Social Media}, 11\penalty0 (1):\penalty0 512--515, 2017.
\newblock \doi{10.1609/icwsm.v11i1.14955}.
\newblock URL \url{https://doi.org/10.1609/icwsm.v11i1.14955}.

\bibitem[DeYoung et~al.(2020)DeYoung, Jain, Rajani, Lehman, Xiong, Socher, and
  Wallace]{deyoung2020eraser}
Jay DeYoung, Sarthak Jain, Nazneen~Fatema Rajani, Eric Lehman, Caiming Xiong,
  Richard Socher, and Byron~C. Wallace.
\newblock {ERASER}: {A} benchmark to evaluate rationalized {NLP} models.
\newblock In \emph{Proceedings of the 58th Annual Meeting of the {A}ssociation
  for {C}omputational {L}inguistics}, pages 4443--4458. Association for
  Computational Linguistics, 2020.
\newblock \doi{10.18653/v1/2020.acl-main.408}.
\newblock URL \url{https://aclanthology.org/2020.acl-main.408/}.

\bibitem[Dieng et~al.(2020)Dieng, Ruiz, and Blei]{dieng2020etm}
Adji~B. Dieng, Francisco J.~R. Ruiz, and David~M. Blei.
\newblock Topic modeling in embedding spaces.
\newblock \emph{Transactions of the Association for Computational Linguistics},
  8:\penalty0 439--453, 2020.
\newblock \doi{10.1162/tacl_a_00325}.
\newblock URL \url{https://aclanthology.org/2020.tacl-1.29/}.

\bibitem[Dixon et~al.(2018)Dixon, Li, Sorensen, Thain, and
  Vasserman]{dixon2018bias}
Lucas Dixon, John Li, Jeffrey Sorensen, Nithum Thain, and Lucy Vasserman.
\newblock Measuring and mitigating unintended bias in text classification.
\newblock In \emph{Proceedings of the 2018 AAAI/ACM Conference on AI, Ethics,
  and Society}, pages 67--73. ACM, 2018.
\newblock \doi{10.1145/3278721.3278729}.
\newblock URL \url{https://doi.org/10.1145/3278721.3278729}.

\bibitem[Dodge et~al.(2019)Dodge, Gururangan, Card, Schwartz, and
  Smith]{dodge2019reporting}
Jesse Dodge, Suchin Gururangan, Dallas Card, Roy Schwartz, and Noah~A. Smith.
\newblock Show your work: Improved reporting of experimental results.
\newblock In \emph{Proceedings of the 2019 Conference on Empirical Methods in
  Natural Language Processing and the 9th International Joint Conference on
  Natural Language Processing (EMNLP-IJCNLP)}, pages 2185--2194. Association
  for Computational Linguistics, 2019.
\newblock \doi{10.18653/v1/D19-1224}.
\newblock URL \url{https://aclanthology.org/D19-1224/}.

\bibitem[Douglas et~al.(2026)Douglas, Balci, and
  Aylett-Bullock]{douglas2026prism}
Connor Douglas, Utkucan Balci, and Joseph Aylett-Bullock.
\newblock {PRISM}: {LLM}-guided semantic clustering for high-precision topics.
\newblock In \emph{Proceedings of the ACM Web Conference 2026}, pages
  8701--8704. ACM, 2026.
\newblock \doi{10.1145/3774904.3792944}.

\bibitem[Dror et~al.(2018)Dror, Baumer, Shlomov, and Reichart]{dror2018testing}
Rotem Dror, Gili Baumer, Segev Shlomov, and Roi Reichart.
\newblock The hitchhiker's guide to testing statistical significance in natural
  language processing.
\newblock In \emph{Proceedings of the 56th Annual Meeting of the Association
  for Computational Linguistics (Volume 1: Long Papers)}, pages 1383--1392.
  Association for Computational Linguistics, 2018.
\newblock \doi{10.18653/v1/P18-1128}.
\newblock URL \url{https://aclanthology.org/P18-1128/}.

\bibitem[Efron and Tibshirani(1993)]{efron1993bootstrap}
Bradley Efron and Robert~J. Tibshirani.
\newblock \emph{An Introduction to the Bootstrap}.
\newblock Chapman \& Hall, 1993.
\newblock ISBN 9780412042317.
\newblock \doi{10.1007/978-1-4899-4541-9}.
\newblock URL \url{https://doi.org/10.1007/978-1-4899-4541-9}.

\bibitem[Eilertsen et~al.(2026)Eilertsen, Bj{\o}rgfinsd{\'o}ttir, Vargas, and
  Ramezani-Kebrya]{eilertsen2026rationales}
Brage Eilertsen, R{\o}skva Bj{\o}rgfinsd{\'o}ttir, Francielle Vargas, and Ali
  Ramezani-Kebrya.
\newblock Aligning attention with human rationales for self-explaining hate
  speech detection.
\newblock \emph{Proceedings of the AAAI Conference on Artificial Intelligence},
  40\penalty0 (44):\penalty0 37369--37378, 2026.
\newblock \doi{10.1609/aaai.v40i44.41069}.

\bibitem[Fan et~al.(2008)Fan, Chang, Hsieh, Wang, and Lin]{fan2008liblinear}
Rong-En Fan, Kai-Wei Chang, Cho-Jui Hsieh, Xiang-Rui Wang, and Chih-Jen Lin.
\newblock {LIBLINEAR}: A library for large linear classification.
\newblock \emph{Journal of Machine Learning Research}, 9\penalty0
  (61):\penalty0 1871--1874, 2008.
\newblock URL \url{https://jmlr.org/papers/v9/fan08a.html}.

\bibitem[Feng et~al.(2026)Feng, Ding, Tan, Ma, Li, Wu, Gao, Zhai, and
  Guo]{feng2026agenthazard}
Yunhao Feng, Yifan Ding, Yingshui Tan, Xingjun Ma, Yige Li, Yutao Wu, Yifeng
  Gao, Kun Zhai, and Yanming Guo.
\newblock {AgentHazard}: A benchmark for evaluating harmful behavior in
  computer-use agents.
\newblock arXiv preprint arXiv:2604.02947, 2026.
\newblock URL \url{https://arxiv.org/abs/2604.02947}.

\bibitem[Grootendorst(2022)]{grootendorst2022bertopic}
Maarten Grootendorst.
\newblock {BERTopic}: Neural topic modeling with a class-based {TF-IDF}
  procedure.
\newblock arXiv:2203.05794, 2022.
\newblock URL \url{https://arxiv.org/abs/2203.05794}.

\bibitem[Hoyle et~al.(2021)Hoyle, Goel, Peskov, Hian-Cheong, Boyd-Graber, and
  Resnik]{hoyle2021coherence}
Alexander Hoyle, Pranav Goel, Denis Peskov, Andrew Hian-Cheong, Jordan
  Boyd-Graber, and Philip Resnik.
\newblock Is automated topic model evaluation broken? the incoherence of
  coherence.
\newblock In \emph{Advances in Neural Information Processing Systems},
  volume~34, pages 2018--2033. Curran Associates, Inc., 2021.
\newblock URL
  \url{https://proceedings.neurips.cc/paper_files/paper/2021/file/0f83556a305d789b1d71815e8ea4f4b0-Paper.pdf}.

\bibitem[Jacovi and Goldberg(2020)]{jacovi2020faithfulness}
Alon Jacovi and Yoav Goldberg.
\newblock Towards faithfully interpretable {NLP} systems: How should we define
  and evaluate faithfulness?
\newblock In \emph{Proceedings of the 58th Annual Meeting of the {A}ssociation
  for {C}omputational {L}inguistics}, pages 4198--4205. Association for
  Computational Linguistics, 2020.
\newblock \doi{10.18653/v1/2020.acl-main.386}.
\newblock URL \url{https://aclanthology.org/2020.acl-main.386/}.

\bibitem[Jain and Wallace(2019)]{jain2019attention}
Sarthak Jain and Byron~C. Wallace.
\newblock {A}ttention is not {E}xplanation.
\newblock In \emph{Proceedings of the 2019 Conference of the {N}orth {A}merican
  Chapter of the {A}ssociation for {C}omputational {L}inguistics: Human
  Language Technologies, Volume 1 (Long and Short Papers)}, pages 3543--3556.
  Association for Computational Linguistics, 2019.
\newblock \doi{10.18653/v1/N19-1357}.
\newblock URL \url{https://aclanthology.org/N19-1357/}.

\bibitem[Jhaver et~al.(2019)Jhaver, Bruckman, and
  Gilbert]{jhaver2019transparency}
Shagun Jhaver, Amy Bruckman, and Eric Gilbert.
\newblock Does transparency in moderation really matter?: User behavior after
  content removal explanations on reddit.
\newblock \emph{Proceedings of the ACM on Human-Computer Interaction},
  3\penalty0 (CSCW), 2019.
\newblock \doi{10.1145/3359252}.
\newblock URL \url{https://doi.org/10.1145/3359252}.

\bibitem[Kennedy et~al.(2020)Kennedy, Jin, Mostafazadeh~Davani, Dehghani, and
  Ren]{kennedy2020contextualizing}
Brendan Kennedy, Xisen Jin, Aida Mostafazadeh~Davani, Morteza Dehghani, and
  Xiang Ren.
\newblock Contextualizing hate speech classifiers with post-hoc explanation.
\newblock In \emph{Proceedings of the 58th Annual Meeting of the {A}ssociation
  for {C}omputational {L}inguistics}, pages 5435--5442. Association for
  Computational Linguistics, 2020.
\newblock \doi{10.18653/v1/2020.acl-main.483}.
\newblock URL \url{https://aclanthology.org/2020.acl-main.483/}.

\bibitem[Lee and Seung(1999)]{lee1999nmf}
Daniel~D. Lee and H.~Sebastian Seung.
\newblock Learning the parts of objects by non-negative matrix factorization.
\newblock \emph{Nature}, 401:\penalty0 788--791, 1999.
\newblock \doi{10.1038/44565}.

\bibitem[Lei et~al.(2016)Lei, Barzilay, and Jaakkola]{lei2016rationalizing}
Tao Lei, Regina Barzilay, and Tommi Jaakkola.
\newblock Rationalizing neural predictions.
\newblock In \emph{Proceedings of the 2016 Conference on Empirical Methods in
  Natural Language Processing}, pages 107--117. Association for Computational
  Linguistics, 2016.
\newblock \doi{10.18653/v1/D16-1011}.
\newblock URL \url{https://aclanthology.org/D16-1011/}.

\bibitem[Li et~al.(2022)Li, Hecht, and Chancellor]{li2022volunteer}
Hanlin Li, Brent Hecht, and Stevie Chancellor.
\newblock Measuring the monetary value of online volunteer work.
\newblock In \emph{Proceedings of the International AAAI Conference on Web and
  Social Media}, volume~16, pages 596--606. AAAI Press, 2022.
\newblock \doi{10.1609/icwsm.v16i1.19318}.
\newblock URL \url{https://doi.org/10.1609/icwsm.v16i1.19318}.

\bibitem[Lin et~al.(2026)Lin, Deng, Li, Ma, Feng, Qing, Li, Zhang, Cui, Meng,
  Zhang, Ma, Li, Xu, and Ji]{lin2026selfevolvingsafety}
Ruixiao Lin, Xinhao Deng, Qingming Li, Jianan Ma, Yunhao Feng, Yuqi Qing,
  Zhenyuan Li, Yechao Zhang, Shiwen Cui, Changhua Meng, Tianwei Zhang, Xingjun
  Ma, Qi~Li, Ke~Xu, and Shouling Ji.
\newblock Safety in self-evolving {LLM} agent systems: Threats, amplification,
  and case studies.
\newblock arXiv preprint arXiv:2606.23075, 2026.
\newblock URL \url{https://arxiv.org/abs/2606.23075}.

\bibitem[Liu et~al.(2023)Liu, Iter, Xu, Wang, Xu, and Zhu]{liu2023geval}
Yang Liu, Dan Iter, Yichong Xu, Shuohang Wang, Ruochen Xu, and Chenguang Zhu.
\newblock {G-Eval}: {NLG} evaluation using {GPT-4} with better human alignment.
\newblock In \emph{Proceedings of the 2023 Conference on Empirical Methods in
  Natural Language Processing}, pages 2511--2522. Association for Computational
  Linguistics, 2023.
\newblock \doi{10.18653/v1/2023.emnlp-main.153}.
\newblock URL \url{https://aclanthology.org/2023.emnlp-main.153/}.

\bibitem[Lu et~al.(2023)Lu, Xu, Zhang, Min, Yang, and Lin]{toxicn2023}
Junyu Lu, Bo~Xu, Xiaokun Zhang, Changrong Min, Liang Yang, and Hongfei Lin.
\newblock Facilitating fine-grained detection of {Chinese} toxic language:
  Hierarchical taxonomy, resources, and benchmarks.
\newblock In \emph{Proceedings of the 61st Annual Meeting of the Association
  for Computational Linguistics (Volume 1: Long Papers)}, pages 16235--16250.
  Association for Computational Linguistics, 2023.
\newblock \doi{10.18653/v1/2023.acl-long.898}.
\newblock URL \url{https://aclanthology.org/2023.acl-long.898/}.

\bibitem[Ma et~al.(2025)Ma, Gao, Wang, Wang, Wang, Sun, Ding, Xu, Chen, Zhao,
  Huang, Li, Wu, Zhang, Zheng, Bai, Li, Wu, Qiu, Zhang, Han, Li, Sun, Wang, Gu,
  Wu, Chen, Zhang, Liu, Gong, Liu, Pan, Xie, Pang, Dong, Jia, Zhang, Ma, Zhang,
  Gong, Xiao, Erfani, Baldwin, Li, Sugiyama, Tao, Bailey, and
  Jiang]{ma2025safetyatscale}
Xingjun Ma, Yifeng Gao, Yixu Wang, Ruofan Wang, Xin Wang, Ye~Sun, Yifan Ding,
  Hengyuan Xu, Yunhao Chen, Yunhao Zhao, Hanxun Huang, Yige Li, Yutao Wu,
  Jiaming Zhang, Xiang Zheng, Yang Bai, Yiming Li, Zuxuan Wu, Xipeng Qiu,
  Jingfeng Zhang, Xudong Han, Haonan Li, Jun Sun, Cong Wang, Jindong Gu,
  Baoyuan Wu, Siheng Chen, Tianwei Zhang, Yang Liu, Mingming Gong, Tongliang
  Liu, Shirui Pan, Cihang Xie, Tianyu Pang, Yinpeng Dong, Ruoxi Jia, Yang
  Zhang, Shiqing Ma, Xiangyu Zhang, Neil Gong, Chaowei Xiao, Sarah Erfani, Tim
  Baldwin, Bo~Li, Masashi Sugiyama, Dacheng Tao, James Bailey, and Yu-Gang
  Jiang.
\newblock Safety at scale: A comprehensive survey of large model and agent
  safety.
\newblock \emph{Foundations and Trends in Privacy and Security}, 8\penalty0
  (3-4):\penalty0 254--469, 2025.
\newblock \doi{10.1561/3300000051}.
\newblock URL \url{https://doi.org/10.1561/3300000051}.

\bibitem[Mathew et~al.(2021)Mathew, Saha, Yimam, Biemann, Goyal, and
  Mukherjee]{mathew2021hatexplain}
Binny Mathew, Punyajoy Saha, Seid~Muhie Yimam, Chris Biemann, Pawan Goyal, and
  Animesh Mukherjee.
\newblock {HateXplain}: A benchmark dataset for explainable hate speech
  detection.
\newblock \emph{Proceedings of the AAAI Conference on Artificial Intelligence},
  35\penalty0 (17):\penalty0 14867--14875, 2021.
\newblock \doi{10.1609/aaai.v35i17.17745}.

\bibitem[Pavlopoulos et~al.(2020)Pavlopoulos, Sorensen, Dixon, Thain, and
  Androutsopoulos]{pavlopoulos2020context}
John Pavlopoulos, Jeffrey Sorensen, Lucas Dixon, Nithum Thain, and Ion
  Androutsopoulos.
\newblock Toxicity detection: Does context really matter?
\newblock In \emph{Proceedings of the 58th Annual Meeting of the {A}ssociation
  for {C}omputational {L}inguistics}, pages 4296--4305. Association for
  Computational Linguistics, 2020.
\newblock \doi{10.18653/v1/2020.acl-main.396}.
\newblock URL \url{https://aclanthology.org/2020.acl-main.396/}.

\bibitem[Pham et~al.(2024)Pham, Hoyle, Sun, Resnik, and
  Iyyer]{pham2024topicgpt}
Chau~Minh Pham, Alexander Hoyle, Simeng Sun, Philip Resnik, and Mohit Iyyer.
\newblock {T}opic{GPT}: A prompt-based topic modeling framework.
\newblock In \emph{Proceedings of the 2024 Conference of the {N}orth {A}merican
  Chapter of the {A}ssociation for {C}omputational {L}inguistics: Human
  Language Technologies (Volume 1: Long Papers)}, pages 2956--2984. Association
  for Computational Linguistics, 2024.
\newblock \doi{10.18653/v1/2024.naacl-long.164}.
\newblock URL \url{https://aclanthology.org/2024.naacl-long.164/}.

\bibitem[Reimers and Gurevych(2019)]{reimers2019sbert}
Nils Reimers and Iryna Gurevych.
\newblock Sentence-{BERT}: Sentence embeddings using {Siamese BERT}-networks.
\newblock In \emph{Proceedings of the 2019 Conference on Empirical Methods in
  Natural Language Processing and the 9th International Joint Conference on
  Natural Language Processing (EMNLP-IJCNLP)}, pages 3982--3992. Association
  for Computational Linguistics, 2019.
\newblock \doi{10.18653/v1/D19-1410}.
\newblock URL \url{https://aclanthology.org/D19-1410/}.

\bibitem[Reimers and Gurevych(2020)]{reimers2020multilingual}
Nils Reimers and Iryna Gurevych.
\newblock Making monolingual sentence embeddings multilingual using knowledge
  distillation.
\newblock In \emph{Proceedings of the 2020 Conference on Empirical Methods in
  Natural Language Processing (EMNLP)}, pages 4512--4525. Association for
  Computational Linguistics, 2020.
\newblock \doi{10.18653/v1/2020.emnlp-main.365}.
\newblock URL \url{https://aclanthology.org/2020.emnlp-main.365/}.

\bibitem[R{\"o}der et~al.(2015)R{\"o}der, Both, and
  Hinneburg]{roder2015coherence}
Michael R{\"o}der, Andreas Both, and Alexander Hinneburg.
\newblock Exploring the space of topic coherence measures.
\newblock In \emph{Proceedings of the Eighth ACM International Conference on
  Web Search and Data Mining}, pages 399--408. ACM, 2015.
\newblock \doi{10.1145/2684822.2685324}.
\newblock URL \url{https://doi.org/10.1145/2684822.2685324}.

\bibitem[R{\"o}ttger et~al.(2021)R{\"o}ttger, Vidgen, Nguyen, Waseem, Margetts,
  and Pierrehumbert]{rottger2021hatecheck}
Paul R{\"o}ttger, Bertie Vidgen, Dong Nguyen, Zeerak Waseem, Helen Margetts,
  and Janet Pierrehumbert.
\newblock {H}ate{C}heck: Functional tests for hate speech detection models.
\newblock In \emph{Proceedings of the 59th Annual Meeting of the {A}ssociation
  for {C}omputational {L}inguistics and the 11th International Joint Conference
  on Natural Language Processing (Volume 1: Long Papers)}, pages 41--58.
  Association for Computational Linguistics, 2021.
\newblock \doi{10.18653/v1/2021.acl-long.4}.
\newblock URL \url{https://aclanthology.org/2021.acl-long.4/}.

\bibitem[Saito and Rehmsmeier(2015)]{saito2015pr}
Takaya Saito and Marc Rehmsmeier.
\newblock The precision-recall plot is more informative than the {ROC} plot
  when evaluating binary classifiers on imbalanced datasets.
\newblock \emph{PLOS ONE}, 10\penalty0 (3):\penalty0 e0118432, 2015.
\newblock \doi{10.1371/journal.pone.0118432}.
\newblock URL
  \url{https://journals.plos.org/plosone/article?id=10.1371/journal.pone.0118432}.

\bibitem[Sap et~al.(2019)Sap, Card, Gabriel, Choi, and Smith]{sap2019racial}
Maarten Sap, Dallas Card, Saadia Gabriel, Yejin Choi, and Noah~A. Smith.
\newblock The risk of racial bias in hate speech detection.
\newblock In \emph{Proceedings of the 57th Annual Meeting of the {A}ssociation
  for {C}omputational {L}inguistics}, pages 1668--1678. Association for
  Computational Linguistics, 2019.
\newblock \doi{10.18653/v1/P19-1163}.
\newblock URL \url{https://aclanthology.org/P19-1163/}.

\bibitem[{ScamSpot}(n.d.)]{scamspotrepo}
{ScamSpot}.
\newblock Instagram data scraper.
\newblock GitHub dataset and code repository, n.d.
\newblock URL \url{https://github.com/ScamSpot/scamspot_ig-comment-scraper}.
\newblock Accessed 17 October 2026.

\bibitem[{Sentence Transformers}(n.d.)]{minilmModel}
{Sentence Transformers}.
\newblock paraphrase-multilingual-{MiniLM-L12-v2}: Model card.
\newblock Hugging Face model repository, n.d.
\newblock URL
  \url{https://huggingface.co/sentence-transformers/paraphrase-multilingual-MiniLM-L12-v2}.
\newblock Accessed 17 October 2026.

\bibitem[Vargas et~al.(2022)Vargas, Carvalho, Rodrigues~de G{\'o}es, Pardo, and
  Benevenuto]{hatebr2022}
Francielle Vargas, Isabelle Carvalho, Fabiana Rodrigues~de G{\'o}es, Thiago
  Pardo, and Fabr{\'i}cio Benevenuto.
\newblock {HateBR}: A large expert annotated corpus of {Brazilian Instagram}
  comments for offensive language and hate speech detection.
\newblock In \emph{Proceedings of the Thirteenth Language Resources and
  Evaluation Conference}, pages 7174--7183. European Language Resources
  Association, 2022.
\newblock URL \url{https://aclanthology.org/2022.lrec-1.777/}.

\bibitem[Vidgen et~al.(2021)Vidgen, Thrush, Waseem, and
  Kiela]{vidgen2021dynamic}
Bertie Vidgen, Tristan Thrush, Zeerak Waseem, and Douwe Kiela.
\newblock Learning from the worst: Dynamically generated datasets to improve
  online hate detection.
\newblock In \emph{Proceedings of the 59th Annual Meeting of the {A}ssociation
  for {C}omputational {L}inguistics and the 11th International Joint Conference
  on Natural Language Processing (Volume 1: Long Papers)}, pages 1667--1682.
  Association for Computational Linguistics, 2021.
\newblock \doi{10.18653/v1/2021.acl-long.132}.
\newblock URL \url{https://aclanthology.org/2021.acl-long.132/}.

\bibitem[Vogels(2021)]{vogels2021harassment}
Emily~A. Vogels.
\newblock The state of online harassment.
\newblock Pew Research Center report, 2021.
\newblock URL
  \url{https://www.pewresearch.org/internet/2021/01/13/the-state-of-online-harassment/}.
\newblock Accessed 26 October 2026.

\bibitem[Wang et~al.(2025)Wang, Zhang, Zhou, Wu, Yu, Zhao, Yin, Fu, Yan, Luo,
  Lin, Xu, Lu, Cao, Zhou, Jin, Meng, Xu, Mao, Wang, Wu, Wang, Zhang, Fang, Qu,
  Liu, Liu, Zhang, Li, Guo, Qin, Fan, Wang, Ding, Hong, Ji, Lai, Yu, Li, Jiang,
  Li, Deng, Wu, Wang, Huang, Guo, tse Huang, Wang, Jin, Wang, Liu, Yue, Huang,
  Wan, Chang, Li, Yu, Li, Li, Bai, Zhang, Guo, Wang, Chen, Zhou, Jia, Sun, Wu,
  Chen, Hu, Li, Wang, Zhang, Tuan, Xu, Zhang, Zhang, Ma, Gu, Pang, Wang, An,
  Sun, Bansal, Pan, Lyu, Elovici, Kailkhura, Yang, Li, Xu, Sun, Wang, Li, Tang,
  Jiang, Juefei-Xu, Xiong, Wang, Tao, Yu, Wen, and
  Liu]{wang2025fullstacksafety}
Kun Wang, Guibin Zhang, Zhenhong Zhou, Jiahao Wu, Miao Yu, Shiqian Zhao,
  Chenlong Yin, Jinhu Fu, Yibo Yan, Hanjun Luo, Liang Lin, Zhihao Xu, Haolang
  Lu, Xinye Cao, Xinyun Zhou, Weifei Jin, Fanci Meng, Shicheng Xu, Junyuan Mao,
  Yu~Wang, Hao Wu, Minghe Wang, Fan Zhang, Junfeng Fang, Wenjie Qu, Yue Liu,
  Chengwei Liu, Yifan Zhang, Qiankun Li, Chongye Guo, Yalan Qin, Zhaoxin Fan,
  Kai Wang, Yi~Ding, Donghai Hong, Jiaming Ji, Yingxin Lai, Zitong Yu, Xinfeng
  Li, Yifan Jiang, Yanhui Li, Xinyu Deng, Junlin Wu, Dongxia Wang, Yihao Huang,
  Yufei Guo, Jen tse Huang, Qiufeng Wang, Xiaolong Jin, Wenxuan Wang, Dongrui
  Liu, Yanwei Yue, Wenke Huang, Guancheng Wan, Heng Chang, Tianlin Li, Yi~Yu,
  Chenghao Li, Jiawei Li, Lei Bai, Jie Zhang, Qing Guo, Jingyi Wang, Tianlong
  Chen, Joey~Tianyi Zhou, Xiaojun Jia, Weisong Sun, Cong Wu, Jing Chen, Xuming
  Hu, Yiming Li, Xiao Wang, Ningyu Zhang, Luu~Anh Tuan, Guowen Xu, Jiaheng
  Zhang, Tianwei Zhang, Xingjun Ma, Jindong Gu, Liang Pang, Xiang Wang, Bo~An,
  Jun Sun, Mohit Bansal, Shirui Pan, Lingjuan Lyu, Yuval Elovici, Bhavya
  Kailkhura, Yaodong Yang, Hongwei Li, Wenyuan Xu, Yizhou Sun, Wei Wang, Qing
  Li, Ke~Tang, Yu-Gang Jiang, Felix Juefei-Xu, Hui Xiong, Xiaofeng Wang,
  Dacheng Tao, Philip~S. Yu, Qingsong Wen, and Yang Liu.
\newblock A comprehensive survey in {LLM}(-agent) full stack safety: Data,
  training and deployment.
\newblock arXiv preprint arXiv:2504.15585, 2025.
\newblock URL \url{https://arxiv.org/abs/2504.15585}.

\bibitem[Wang et~al.(2024)Wang, Li, Chen, Cai, Zhu, Lin, Cao, Kong, Liu, Liu,
  and Sui]{wang2024fair}
Peiyi Wang, Lei Li, Liang Chen, Zefan Cai, Dawei Zhu, Binghuai Lin, Yunbo Cao,
  Lingpeng Kong, Qi~Liu, Tianyu Liu, and Zhifang Sui.
\newblock Large language models are not fair evaluators.
\newblock In \emph{Proceedings of the 62nd Annual Meeting of the Association
  for Computational Linguistics (Volume 1: Long Papers)}, pages 9440--9450.
  Association for Computational Linguistics, 2024.
\newblock \doi{10.18653/v1/2024.acl-long.511}.
\newblock URL \url{https://aclanthology.org/2024.acl-long.511/}.

\bibitem[Wang et~al.(2020)Wang, Wei, Dong, Bao, Yang, and Zhou]{wang2020minilm}
Wenhui Wang, Furu Wei, Li~Dong, Hangbo Bao, Nan Yang, and Ming Zhou.
\newblock {MiniLM}: Deep self-attention distillation for task-agnostic
  compression of pre-trained transformers.
\newblock In \emph{Advances in Neural Information Processing Systems},
  volume~33, pages 5776--5788. Curran Associates, Inc., 2020.
\newblock URL
  \url{https://proceedings.neurips.cc/paper_files/paper/2020/file/3f5ee243547dee91fbd053c1c4a845aa-Paper.pdf}.

\bibitem[Weinberger et~al.(2009)Weinberger, Dasgupta, Langford, Smola, and
  Attenberg]{weinberger2009hashing}
Kilian Weinberger, Anirban Dasgupta, John Langford, Alex Smola, and Josh
  Attenberg.
\newblock Feature hashing for large scale multitask learning.
\newblock In \emph{Proceedings of the 26th Annual International Conference on
  Machine Learning}, pages 1113--1120. ACM, 2009.
\newblock \doi{10.1145/1553374.1553516}.
\newblock URL \url{https://doi.org/10.1145/1553374.1553516}.

\bibitem[Weld et~al.(2021)Weld, Huang, Lee, Zhang, Wang, Guo, Long, Poon, and
  Han]{conda2021}
Henry Weld, Guanghao Huang, Jean Lee, Tongshu Zhang, Kunze Wang, Xinghong Guo,
  Siqu Long, Josiah Poon, and Caren Han.
\newblock {CONDA}: a {CON}textual dual-annotated dataset for in-game toxicity
  understanding and detection.
\newblock In \emph{Findings of the Association for Computational Linguistics:
  ACL-IJCNLP 2021}, pages 2406--2416. Association for Computational
  Linguistics, 2021.
\newblock \doi{10.18653/v1/2021.findings-acl.213}.
\newblock URL \url{https://aclanthology.org/2021.findings-acl.213/}.

\bibitem[Wulczyn et~al.(2017)Wulczyn, Thain, and Dixon]{wulczyn2017exmachina}
Ellery Wulczyn, Nithum Thain, and Lucas Dixon.
\newblock Ex machina: Personal attacks seen at scale.
\newblock In \emph{Proceedings of the 26th International Conference on World
  Wide Web}, pages 1391--1399. International World Wide Web Conferences
  Steering Committee, 2017.
\newblock \doi{10.1145/3038912.3052591}.
\newblock URL \url{https://doi.org/10.1145/3038912.3052591}.

\bibitem[Yan et~al.(2013)Yan, Guo, Lan, and Cheng]{yan2013btm}
Xiaohui Yan, Jiafeng Guo, Yanyan Lan, and Xueqi Cheng.
\newblock A biterm topic model for short texts.
\newblock In \emph{Proceedings of the 22nd International Conference on World
  Wide Web}, pages 1445--1456. ACM, 2013.
\newblock \doi{10.1145/2488388.2488514}.
\newblock URL \url{https://doi.org/10.1145/2488388.2488514}.

\bibitem[Yang et~al.(2025)Yang, Chen, Xu, Shang, Guo, and Zhang]{sccd2025}
Qingpo Yang, Yakai Chen, Zihui Xu, Yu-ming Shang, Sanchuan Guo, and Xi~Zhang.
\newblock {SCCD}: A session-based dataset for {Chinese} cyberbullying
  detection.
\newblock In \emph{Proceedings of the 31st International Conference on
  Computational Linguistics}, pages 9533--9545. Association for Computational
  Linguistics, 2025.
\newblock URL \url{https://aclanthology.org/2025.coling-main.639/}.

\bibitem[Yin and Wang(2014)]{yin2014gsdmm}
Jianhua Yin and Jianyong Wang.
\newblock A {Dirichlet} multinomial mixture model-based approach for short text
  clustering.
\newblock In \emph{Proceedings of the 20th ACM SIGKDD International Conference
  on Knowledge Discovery and Data Mining}, pages 233--242. ACM, 2014.
\newblock \doi{10.1145/2623330.2623715}.
\newblock URL \url{https://doi.org/10.1145/2623330.2623715}.

\bibitem[Yu et~al.(2022)Yu, Blanco, and Hong]{yu2022context}
Xinchen Yu, Eduardo Blanco, and Lingzi Hong.
\newblock Hate speech and counter speech detection: Conversational context does
  matter.
\newblock In \emph{Proceedings of the 2022 Conference of the {N}orth {A}merican
  Chapter of the {A}ssociation for {C}omputational {L}inguistics: Human
  Language Technologies}, pages 5918--5930. Association for Computational
  Linguistics, 2022.
\newblock \doi{10.18653/v1/2022.naacl-main.433}.
\newblock URL \url{https://aclanthology.org/2022.naacl-main.433/}.

\bibitem[Zhang et~al.(2018)Zhang, Chang, Danescu-Niculescu-Mizil, Dixon, Hua,
  Taraborelli, and Thain]{zhang2018awry}
Justine Zhang, Jonathan Chang, Cristian Danescu-Niculescu-Mizil, Lucas Dixon,
  Yiqing Hua, Dario Taraborelli, and Nithum Thain.
\newblock Conversations gone awry: Detecting early signs of conversational
  failure.
\newblock In \emph{Proceedings of the 56th Annual Meeting of the {A}ssociation
  for {C}omputational {L}inguistics (Volume 1: Long Papers)}, pages 1350--1361.
  Association for Computational Linguistics, 2018.
\newblock \doi{10.18653/v1/P18-1125}.
\newblock URL \url{https://aclanthology.org/P18-1125/}.

\bibitem[Zhang et~al.(2023)Zhang, Nanduri, Jiang, Wu, and Sap]{zhang2023biasx}
Yiming Zhang, Sravani Nanduri, Liwei Jiang, Tongshuang Wu, and Maarten Sap.
\newblock {B}ias{X}: ``thinking slow'' in toxic content moderation with
  explanations of implied social biases.
\newblock In \emph{Proceedings of the 2023 Conference on Empirical Methods in
  Natural Language Processing}, pages 4920--4932. Association for Computational
  Linguistics, 2023.
\newblock \doi{10.18653/v1/2023.emnlp-main.300}.
\newblock URL \url{https://aclanthology.org/2023.emnlp-main.300/}.

\bibitem[Zheng et~al.(2023)Zheng, Chiang, Sheng, Zhuang, Wu, Zhuang, Lin, Li,
  Li, Xing, Zhang, Gonzalez, and Stoica]{zheng2023judge}
Lianmin Zheng, Wei-Lin Chiang, Ying Sheng, Siyuan Zhuang, Zhanghao Wu, Yonghao
  Zhuang, Zi~Lin, Zhuohan Li, Dacheng Li, Eric Xing, Hao Zhang, Joseph
  Gonzalez, and Ion Stoica.
\newblock Judging {LLM}-as-a-judge with {MT-Bench} and {Chatbot Arena}.
\newblock In \emph{Advances in Neural Information Processing Systems},
  volume~36, pages 46595--46623. Curran Associates, Inc., 2023.
\newblock \doi{10.52202/075280-2020}.
\newblock URL
  \url{https://proceedings.neurips.cc/paper_files/paper/2023/file/91f18a1287b398d378ef22505bf41832-Paper-Datasets_and_Benchmarks.pdf}.

\bibitem[Zheng et~al.(2026)Zheng, Wu, Huang, Li, Ma, Li, Jiang, and
  Wang]{zheng2026justask}
Xiang Zheng, Yutao Wu, Hanxun Huang, Yige Li, Xingjun Ma, Bo~Li, Yu-Gang Jiang,
  and Cong Wang.
\newblock Just ask: Curious code agents reveal system prompts in frontier
  {LLMs}.
\newblock In \emph{Proceedings of the 43rd International Conference on Machine
  Learning}, volume 306 of \emph{Proceedings of Machine Learning Research},
  pages 163714--163738. PMLR, 2026.
\newblock URL \url{https://proceedings.mlr.press/v306/zheng26w.html}.

\end{thebibliography}
\FloatBarrier
\beginappendix
\setlength{\textfloatsep}{10pt plus 2pt minus 2pt}
\setlength{\floatsep}{10pt plus 2pt minus 2pt}
\section{Implementation and Partial Supervision}
\label{app:implementation}

\paragraph{\normalfont\bfseries Records and masks.}
The 83,295 records include one SCCD root-post fallback excluded from message counts, head fitting, and PCA. The 642 unlabeled training messages contribute only representation statistics. Heads use 47,516 attack, 7,147 hate, and 1,847 spam/scam labels, with ToxiCN attack--hate overlap. Features exclude source names, class IDs, and collection metadata.

\paragraph{\normalfont\bfseries Encoding and weights.}
We use \texttt{paraphrase-\allowbreak multilingual-\allowbreak MiniLM-\allowbreak L12-v2}. For $T$ content tokens, 112-token blocks overlap by 16, giving $J=1+\lceil\max(T-112,0)/96\rceil$ blocks; empty text retains one block. With $\operatorname{norm}_2(v)=v/\|v\|_2$ for nonzero $v$,
\begin{equation}
e_{it}=\operatorname{norm}_2\!\left(J_{it}^{-1}\sum_{j=1}^{J_{it}}E(x_{it}^{(j)})\right).
\label{eq:encoding}
\end{equation}
Character hashing preserves case, uses L2 normalization, and fits no vocabulary. For category $c$, let $N_c$ count observed labels, $S_c$ contributing sources, $K_{sc}$ observed classes in source $s$, and $N_{scy}$ their counts:
\begin{equation}
w_{itc}=\frac{N_c}{S_c K_{sc}N_{scy_{itc}}}.
\label{eq:weights}
\end{equation}
Weights average one. LIBLINEAR uses L2 regularization, $C=1$, at most 1,000 iterations, and a regularized intercept with unit scaling.

\paragraph{\normalfont\bfseries Discussion summaries.}
The ten risk features are
\begin{equation}
\begin{split}
z_i^{\mathrm{risk}}=[&\mu(a),\max(a),\sigma(a),Q_{.25}(a),Q_{.50}(a),\\
&Q_{.75}(a),Q_{.90}(a),\mu_{\mathrm{top}20\%}(a),\\
&n_i^{-1}\textstyle\sum_t\mathbf{1}(a_{it}\geq0.5),\log(1+n_i)].
\end{split}
\end{equation}
SD uses the population denominator; quantiles use linear interpolation; the upper tail contains $\max(1,\lceil0.2n_i\rceil)$ messages. PCA uses seeded randomized projections. For parent edge $(t,p)$, $d^e=1-\cos(e_{it},e_{ip})$ and $d^a=|a_{it}-a_{ip}|$. Reply features are mean/SD/max of $d^e$, mean/max of $d^a$, and valid-edge fraction. Empty summaries are zero. Threshold ties favor the lower value.

\begin{figure}[t]
\centering
\includegraphics[width=0.62\linewidth]{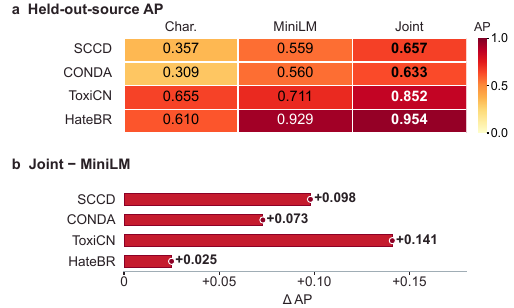}
\caption{Cross-source attack AP and joint-minus-MiniLM gains.}
\Description{A heatmap compares three representations on SCCD, CONDA, ToxiCN, and HateBR. The joint representation has AP 0.657, 0.633, 0.852, and 0.954. A second panel shows positive joint-minus-MiniLM differences of 0.098, 0.073, 0.141, and 0.025. No seed standard deviations or confidence intervals are displayed.}
\label{fig:stability-transfer}
\end{figure}

\paragraph{\normalfont\bfseries Extractive grouping.}
Complete-link clustering uses threshold 0.35 and distance $\operatorname{clip}(1-e_{iu}^{\mathsf T}e_{iv},0,2)$. It becomes one when exactly one message matches counter-speech patterns and the other reaches its attack threshold. Names exclude generic phrases and use one to four tokens, capped at 12 Chinese or 36 other characters. Negated or interrogative clauses remain whole up to 96 non-whitespace characters. For phrase $\omega$, support count $f_{k\omega}$, group size $n_k$, group count $df_\omega$, and $K$ groups,
\begin{equation}
\begin{split}
Q_{k\omega}={}&\frac{f_{k\omega}}{n_k}\log\!\left(1+\frac{K+1}{df_\omega+1}\right)\\
&\times\left[1+0.12\min(\ell_\omega,8)\right],
\end{split}
\label{eq:naming}
\end{equation}
where $\ell_\omega$ excludes ordinary spaces. Ranking and substring suppression are deterministic. Excerpts retain spelling and half-open Unicode offsets. Fragments overlapping protected clauses are discarded; overlength clauses yield no replacement. Parent association requires nameless replies of at most 12 non-whitespace characters.

\paragraph{\normalfont\bfseries Reference fitting and metrics.}
XLM-R uses three epochs, AdamW at $2\times10^{-5}$, weight decay 0.01, 10\% warm-up, and batch size 32. Its 510-token blocks overlap by 64; block logits average before the loss. AP is $\sum_k P_k(R_k-R_{k-1})$ over distinct descending thresholds, with ties entering together and $R_0=0$. ROC-AUC counts positive--negative orderings with half credit for ties. Within repartitions, any calibration uses validation data. Confirmation outcomes inform neither hyperparameters nor evaluation stopping.

\paragraph{\normalfont\bfseries CPU scope.}
We use four threads on AMD64 Family~25 Model~97 with 15.18\,GiB RAM. Encoder size is 499,561,295 bytes (476.419\,MiB). Heads and summaries add 1,494,648 bytes (1.425\,MiB), totaling 477.844\,MiB. Tokenizers, configurations, and runtime dependencies are additional assets. Startup uses a fresh interpreter without OS page-cache eviction and records process/cache state.

\section{Budget Accounting and Uncertainty}
\label{app:budget-accounting}

\paragraph{\normalfont\bfseries Selection and display.}
Discussion ties follow identifiers; message ties follow source order, then identifiers. Selectors take the first $\min(B,n_i)$ messages per top-$L$ discussion. MMR uses
\begin{equation}
\begin{split}
\arg\max_{t\notin S}\,[0.7a_{it}-0.3\rho(t,S)],\\
\rho(t,S)=\max\bigl(\{0\}\cup\{\cos(e_{it},e_{iu}):u\in S\}\bigr).
\end{split}
\label{eq:mmr}
\end{equation}
Display deduplicates identifiers. Character cost includes titles, evidence, added parents, headings, spaces, and punctuation. Character-capped diagnostics follow selector order, skipping candidates whose recomputed cost exceeds the cap without re-ranking. Selected-evidence and displayed-message recall use separate numerators.

\paragraph{\normalfont\bfseries Dependencies and resampling.}
Duplicate links use NFKC normalization, case folding, whitespace removal, and equality or character-5-gram Jaccard $\geq0.85$; texts shorter than five characters require equality.

Paired bootstrap repeats discussions by sampled-cluster multiplicity. Each fitted seed reranks before selecting $L$ copies; recall retains the full dangerous-message denominator. Conditional on fitted models, ordinary percentile endpoints are $0.025$ and $0.975$; family-36 endpoints are $0.025/36$ and $1-0.025/36$, without cross-metric joint coverage. Evidence-only recall is bounded by $\sum_i\min(B,p_i)/\sum_i p_i$, where $p_i$ counts discussion $i$'s positives.

\section{Complete Controls and Sensitivity}
\label{app:additional-results}
\begin{table}[t]
\centering\fontsize{9pt}{11pt}\selectfont
\caption{SCCD component removals. Each cell gives AP after removal above the complete-minus-removal difference. Complete-model message/discussion AP is 0.712/0.976.}
\label{tab:ablations}
\setlength{\tabcolsep}{2.5pt}
\renewcommand{\arraystretch}{1}
\begin{tabularx}{\columnwidth}{@{}>{\raggedright\arraybackslash}p{.28\columnwidth}*{2}{>{\centering\arraybackslash}X}@{}}
\toprule
\rowcolor{CSTTableHeader}
\textbf{Removed} & \textbf{Message AP ($\Delta$)} & \textbf{Discussion AP ($\Delta$)} \\
\midrule
\shortstack[l]{Character\\features} & \shortstack{0.597\\[-2pt](\cstDeltaPos{+0.115})} & \shortstack{0.957\\[-2pt](\cstDeltaPos{+0.019})} \\
\rowcolor{CSTTableStripe}
\shortstack[l]{Message\\semantics} & \shortstack{0.524\\[-2pt](\cstDeltaPos{+0.188})} & \shortstack{0.953\\[-2pt](\cstDeltaPos{+0.023})} \\
\shortstack[l]{Discussion\\semantics} & \shortstack{0.712\\[-2pt](\cstDeltaZero{0})} & \shortstack{0.962\\[-2pt](\cstDeltaPos{+0.014})} \\
\rowcolor{CSTTableStripe}
\shortstack[l]{Reply\\statistics} & \shortstack{0.712\\[-2pt](\cstDeltaZero{0})} & \shortstack{0.959\\[-2pt](\cstDeltaPos{+0.017})} \\
\shortstack[l]{Risk\\statistics} & \shortstack{0.712\\[-2pt](\cstDeltaZero{0})} & \shortstack{0.943\\[-2pt](\cstDeltaPos{+0.033})} \\
\rowcolor{CSTTableStripe}
\shortstack[l]{Source\\balance} & \shortstack{0.686\\[-2pt](\cstDeltaPos{+0.026})} & \shortstack{0.965\\[-2pt](\cstDeltaPos{+0.011})} \\
\bottomrule
\end{tabularx}
\end{table}

\paragraph{\normalfont\bfseries Components and transfer.}
Discussion removals leave message AP unchanged because these features act downstream of its scorer (Table~\ref{tab:ablations}). Source and language co-vary in transfer (Figure~\ref{fig:stability-transfer}).

\section{Judge Rubric and Evaluation Design}
\label{app:judge-rubric}
\begin{table}[t]
\centering\small
\caption{Model-evaluation coverage and agreement. Counts: independent + repeated discussions.}
\label{tab:judge-coverage}
\setlength{\tabcolsep}{3pt}
\begin{tabularx}{\columnwidth}{@{}Xrr@{}}
\toprule
\rowcolor{CSTTableHeader}\textbf{Measure} & \textbf{Gemini} & \textbf{GPT-4o-mini} \\
\midrule
Raw-label discussions & $\textcolor{CSTSrcsccd}{40}\textcolor{CSTSrcconda}{+7}$ & $\textcolor{CSTSrcsccd}{40}\textcolor{CSTSrcconda}{+7}$ \\
\rowcolor{CSTTableStripe}Configuration discussions & $\textcolor{CSTSrcsccd}{40}\textcolor{CSTSrcconda}{+7}$ & $\textcolor{CSTSrcsccd}{40}\textcolor{CSTSrcconda}{+7}$ \\
Fixed-evidence discussions & $\textcolor{CSTSrcsccd}{40}\textcolor{CSTSrcconda}{+7}$ & $\textcolor{CSTSrcsccd}{40}\textcolor{CSTSrcconda}{+7}$ \\
First-attempt validity & $\textcolor{CSTSrchatebr}{\geq183/186}$ & $\textcolor{CSTSrchatebr}{\geq184/186}$ \\
\rowcolor{CSTTableStripe}Percentage (rounded) & \textcolor{CSTSrchatebr}{98.387} & \textcolor{CSTSrchatebr}{98.925} \\
Final response validity (\%) & \textcolor{CSTSrchatebr}{100} & \textcolor{CSTSrchatebr}{100} \\
Repeat agreement, each dimension & $\textcolor{CSTSrchatebr}{7/7}$ & $\textcolor{CSTSrchatebr}{7/7}$ \\
\midrule
\multicolumn{3}{@{}l}{\textbf{Cross-judge raw-label agreement} ($n=38$ jointly known)} \\
Positive-pair Jaccard & \multicolumn{2}{c}{\textcolor{CSTSrchatebr}{0.713}} \\
\rowcolor{CSTTableStripe}All-pair agreement & \multicolumn{2}{c}{\textcolor{CSTSrchatebr}{0.817}} \\
Exact act-set agreement & \multicolumn{2}{c}{\textcolor{CSTSrchatebr}{0.863}} \\
\rowcolor{CSTTableStripe}Act-set Jaccard & \multicolumn{2}{c}{\textcolor{CSTSrchatebr}{0.887}} \\
\bottomrule
\end{tabularx}
\end{table}

\begin{table}[t]
\centering\fontsize{9pt}{11pt}\selectfont
\caption{Additional evaluations and SCCD decision counts.}
\label{tab:extended-evaluation}
\setlength{\tabcolsep}{3pt}
\renewcommand{\arraystretch}{1.05}
\begin{tabularx}{\columnwidth}{@{}>{\raggedright\arraybackslash}p{0.27\columnwidth}X@{}}
\toprule
\rowcolor{CSTTableHeader} Dimension & Observed result and scope \\
\midrule
\cellcolor[HTML]{9DBFE2}\textbf{Development repartitioning} & 10 cluster repartitions; Taskwise AP SD $\leq0.009$; $\geq9/10$ paired wins over MiniLM. Mean AP meets 0.712, 0.887, 0.986, 0.893, 0.879, 0.981. \\
\cellcolor[HTML]{93C9C3}\textbf{SCCD threshold} & Precision 0.719; recall 0.713; macro-F1 0.812. Counts: 8,028 messages, 1,960 positives. \\
\cellcolor[HTML]{BDA5D4}\textbf{External discussions} & $\geq2$ sources; $\geq200$ independent discussions each. Per-source AP $\geq0.913$; paired gain over MiniLM $\geq0.023$. \\
\cellcolor[HTML]{E8BD7A}\textbf{Startup and assets} & Loading $\leq9.713$\,s; first request $\leq1.837$\,s; matched hardware/cache. Encoder and learned assets: 477.844\,MiB; other assets additional. \\
\cellcolor[HTML]{ADD09E}\textbf{Human review} & Median paired time ratio $\leq0.727$ ($\geq27.3\%$ reduction); accuracy $\geq0.947$; missed risk $\leq0.053$; location success $\geq0.973$. \\
\cellcolor[HTML]{D9A6BE}\textbf{Case audit} & 3 platforms; 12 independent discussions each. All tested identifier/span/parent mappings correct; semantic passes $\geq33/36$ (0.917). \\
\cellcolor[HTML]{A8BECD}\textbf{Confirmatory comparisons} & Designated paired AP/evidence-recall difference intervals have positive lower endpoints after family adjustment, using paired predictions and dependency clusters. \\
\bottomrule
\end{tabularx}
\par\smallskip
\begin{tabular*}{\columnwidth}{@{\extracolsep{\fill}}lrrr@{}}
\toprule
Gold & Pred.\ positive & Pred.\ negative & Total \\
\midrule
Positive & 1,397 & 563 & 1,960 \\
Negative & 547 & 5,521 & 6,068 \\
Total & 1,944 & 6,084 & 8,028 \\
\bottomrule
\end{tabular*}
\end{table}

\paragraph{\normalfont\bfseries Rubric.}
Naming asks whether names express source-supported meaning; consistency, whether members share a subject; coverage, whether substantive messages, including ungrouped ones, are represented; stance, whether attack, opposition, quotation, and neutrality are preserved; support, whether cited messages substantiate the output.

\paragraph{\normalfont\bfseries Instructions.}
Chinese prompts score anonymous candidates against source discussions without gold labels: 0 (substantive error), 1 (partly correct), 2 (adequate support), or unknown. Opposing stances within one topic preserve consistency. Known scores require reasons and locally validated message identifiers. Both judges use temperature zero, fresh contexts, and disabled tools; Gemini uses JSON MIME and GPT-4o-mini strict JSON Schema.

\paragraph{\normalfont\bfseries Paired comparisons.}
Figure~\ref{fig:judges} uses 35 jointly scoreable cases for semantic, no-parent, and no-counter-speech contrasts, 32 for BERTopic, and 34 for frequency naming and no-polarity. Means equal summed paired differences divided by jointly known counts, e.g., $8/35=0.229$ after rounding. Failures, output changes, and requests are counted separately. A control whose output is identical to the complete module shares its assessment and is reported as unchanged.

Fixed-evidence comparisons use 36 jointly scoreable cases. Clarity, stance, and traceability difference sums are 8, 3, and 5 for Gemini and 10, 4, and 7 for GPT-4o-mini; division by 36 gives the means. Identical displays contribute zero.

\paragraph{\normalfont\bfseries Validity and agreement.}
Validity fractions in Table~\ref{tab:judge-coverage} express rates, not request totals. Valid responses may include unknowns; attempts and deduplicated requests are counted separately. Full-configuration repeats agree on every dimension ($7/7$).

\paragraph{\normalfont\bfseries Human semantic verification.}
For the panel in Appendix~\ref{app:web-illustration}, assessors independently compare anonymous outputs with sources for name meaning, stance, and claim--evidence support. Agreement is recorded before adjudication; all three checks must pass. Identifier and source-offset checks are separate (Table~\ref{tab:extended-evaluation}).

\paragraph{\normalfont\bfseries Human review.}
Human review follows Section~\ref{sec:protocol}. Time uses the median within-reviewer paired ratio and includes incorrect responses. Against independently adjudicated labels, accuracy covers all assigned binary decisions; missed-risk rate covers gold-positive decisions. Location success requires a correct message identifier and supporting span over all assigned queries. These denominators are separate from semantic cases and model judgments. Paired resampling preserves matched discussion pairs and repeated observations by reviewer and pair. Pilot variance and prespecified effects determine participant and task counts before the main study.

\section{Source-linked Review and Additional Evaluation}
\label{app:web-illustration}
\label{app:extended-evaluation}
\begin{figure}[!htbp]
\centering
\includegraphics[width=0.62\linewidth]{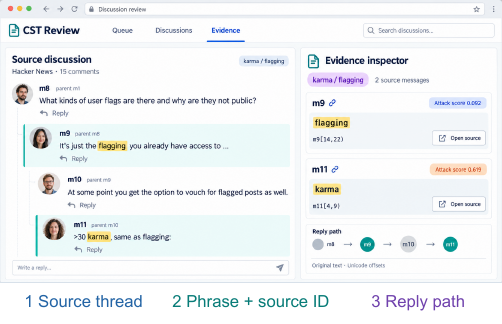}
\caption{Source-linked evidence inspection and reply tracing.}
\Description{A Web discussion interface with source messages on the left and an evidence inspector on the right. Highlighted phrases link to messages m9 and m11; the reply path preserves m10 between them.}
\label{fig:web-review}
\end{figure}

Figure~\ref{fig:web-review} uses Figure~\ref{fig:framework}'s off-benchmark Hacker News snapshot: 15 comments, 14 parent edges, and two deleted items excluded. Without retraining, seed 42 groups $m_9$ and $m_{11}$ as \emph{karma / flagging}, with attack scores 0.092 and 0.619. Spans $m_9[14,22)$ and $m_{11}[4,9)$ recover the name; $m_8\rightarrow m_9\rightarrow m_{10}\rightarrow m_{11}$ retains the intervening reply. Highlighting converts Unicode code-point offsets to JavaScript UTF-16 positions.

\paragraph{\normalfont\bfseries Cross-platform audit.}
The audit fixes platform, discussion-length, and stance-complexity strata before inspecting outputs and excludes training/development duplicates. Identifier, span, and parent-link recovery is checked separately from semantic support (Appendix~\ref{app:judge-rubric}; Table~\ref{tab:extended-evaluation}), with the illustration's attack scores unchanged.

\paragraph{\normalfont\bfseries Additional evaluation.}
Table~\ref{tab:extended-evaluation} follows Table~\ref{tab:message-ap}'s task order. Precision is $1397/1944\approx0.718621$, recall $1397/1960\approx0.712755$, and macro-F1 0.812167, averaging both class F1s. External sources fix risk definitions and source/duplicate identifiers and record prevalence; pilots set sample sizes before testing. Rates use exact counts or bounds before rounding; each evaluation retains its own sampling unit and corresponding denominator.

\end{document}